\documentclass[lettersize,journal]{IEEEtran}
\usepackage{graphicx}
\usepackage{booktabs} %
\usepackage{times}
\usepackage{soul}
\usepackage{url}
\usepackage[hidelinks]{hyperref}
\usepackage{algorithm}
\usepackage{algorithmic}
\usepackage[switch]{lineno}
\usepackage{multirow}
\usepackage{subcaption}
\usepackage{amsmath}
\usepackage{amssymb}
\usepackage{mathtools}
\usepackage{amsthm}
\usepackage[table,xcdraw]{xcolor}
\usepackage{tabularx}
\usepackage{comment}
\usepackage{amsmath}
\usepackage{pifont}

\begin{document}

\title{scDD: Latent Codes Based scRNA-seq Dataset Distillation with Foundation Model Knowledge}

\author{Zhen Yu, Jianan Han, Yang Liu, Qingchao Chen \textsuperscript{\ding{41}}
\thanks{Zhen Yu and Qingchao Chen are from the National Institute of Health Data Science, Peking University, and also with the Institute of Medical Technology, Peking University Health Science Center, Beijing, China, and the State Key Laboratory of General Artificial Intelligence, Peking University, Beijing, China.
Jianan Han is from the AI Research Institute, China Mobile Communications Corporation, Beijing, China.
Yang Liu is from the Wangxuan Institute of Computer Technology, Peking University, Beijing, China.
}
\thanks{\ding{41} Corresponding author. qingchao.chen@pku.edu.cn}
}

\markboth{Journal of \LaTeX\ Class Files,~Vol.~18, No.~9, September~2020}%
{How to Use the IEEEtran \LaTeX \ Templates}

\maketitle

\begin{abstract}

Single-cell RNA sequencing (scRNA-seq) technology has profiled hundreds of millions of human cells across organs, diseases, development and perturbations to date.
However, the high-dimensional sparsity, batch effect noise, category imbalance, and ever-increasing data scale of the original sequencing data pose significant challenges for multi-center knowledge transfer, data fusion, and cross-validation between scRNA-seq datasets.
To address these barriers, (1) we first propose a latent codes-based scRNA-seq dataset distillation framework named scDD, which transfers and distills foundation model knowledge and original dataset information into a compact latent space and generates synthetic scRNA-seq dataset by a generator to replace the original dataset.
Then, (2) we propose a single-step conditional diffusion generator named SCDG, which perform single-step gradient back-propagation to help scDD optimize distillation quality and avoid gradient decay caused by multi-step back-propagation. Meanwhile, SCDG ensures the scRNA-seq data characteristics and inter-class discriminability of the synthetic dataset through flexible conditional control and generation quality assurance.
Finally, we propose a comprehensive benchmark to evaluate the performance of scRNA-seq dataset distillation in different data analysis tasks. It is validated that our proposed method can achieve $7.61\%$ absolute and $15.70\%$ relative improvement over previous state-of-the-art methods on average task.

\end{abstract}

\begin{IEEEkeywords}
scRNA-seq technology, dataset distillation, knowledge transfer and compression, conditional diffusion model, single-cell type annotation.
\end{IEEEkeywords}

\section{Introduction}

\IEEEPARstart{S}ingle-cell RNA sequencing (scRNA-seq) technology~\cite{ahlmann2023comparison,van2023applications,wang2023evolution,chang2024single}, by capturing the detailed gene expression characterizations at the individual cell level, has emerged as a crucial tool for understanding the molecular basis of biological systems and exploring gene expression dynamics in both health and disease. 
However, scRNA-seq data~\cite{maan2024characterizing,yu2024assessing,jiang2022statistics} are inherently high-dimensional and sparse (\textit{e.g.,} with tens thousands dimensions and approximately $90\%$ zero entries), compounded by technical noise and batch effects (\textit{e.g.,} variations in sequencing depth, amplification biases and platform-specific artifacts), extremely imbalanced categories (\textit{e.g.,} the ratio of common cell-types to rare cell-types can reach $70$:$1$), and the ever-growing of modern datasets (\textit{e.g.,} exceeding millions of cells) all poses significant challenges for multi-center knowledge transfer, data fusion and cross verification between scRNA-seq datasets.

\begin{figure}[t]
        \centering
        %\subcaptionbox{Single-step conditional diffusion generator training}
        {
        \includegraphics[width=3.658in]{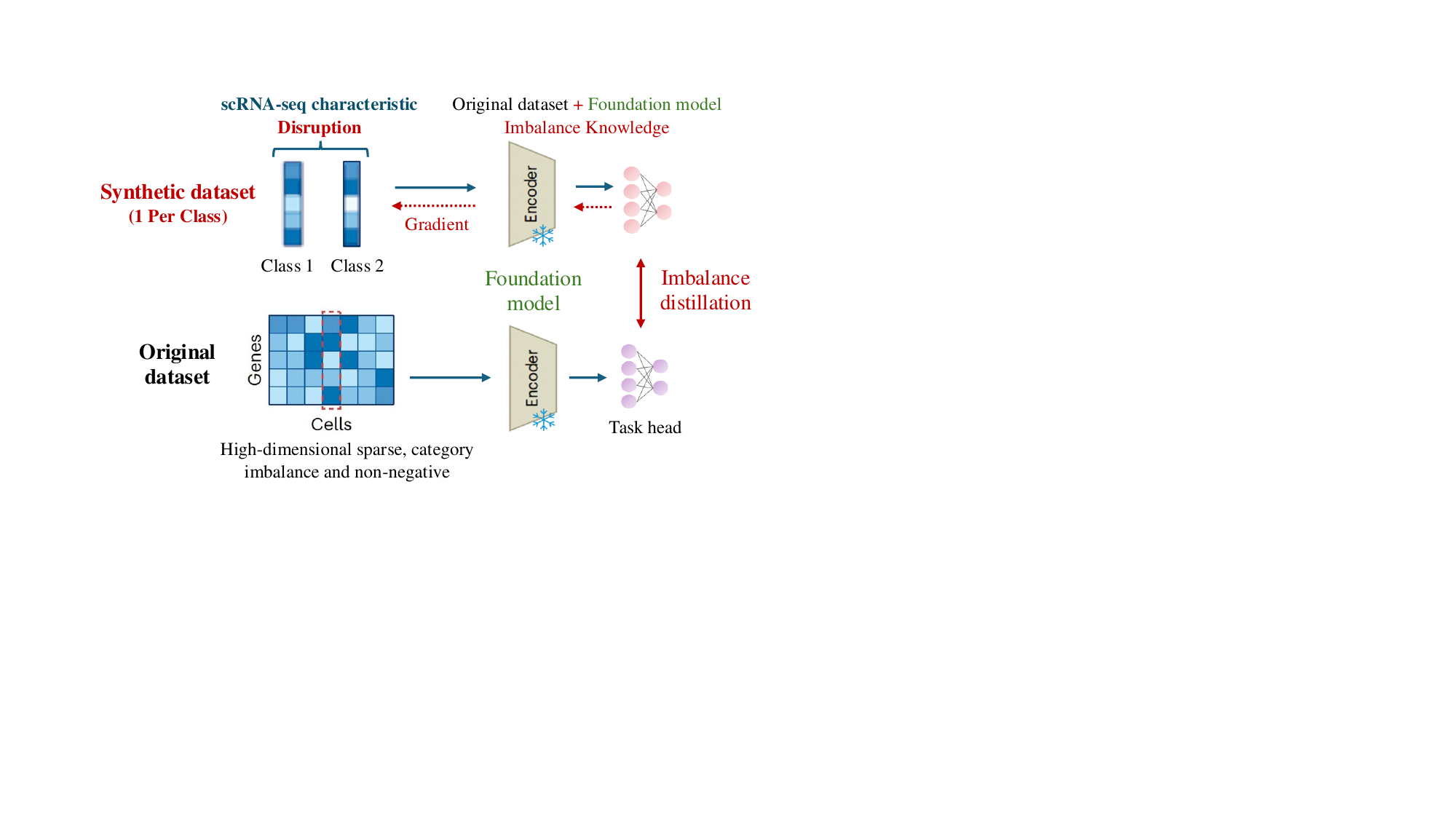}
    }
    \caption{scRNA-seq dataset distillation challenge. (a) directly updating gene expression values at the scRNA-seq data-level will cause them to loss inherently characteristics. (b) distillation with highly categorical imbalance problem leads to a loss of inter-class  discriminability.}
    \label{tensor}
\end{figure}

Dataset distillation~\cite{geng2023survey,lei2023comprehensive2,sachdeva2023data,yu2026dataset} seems a good candidate to solve previous barriers, which can condense a redundant original scRNA-seq dataset into a compact synthetic one, preserving the useful, discriminative and complete information from the original dataset. 
Meanwhile training on these synthetic ones is expected to achieve similar performance compared with training on the original dataset.
\textit{i.e.,} these distilled small-scale synthetic scRNA-seq datasets retain the true and compact biological information from the large-scale, noisy, high-dimensional, category imbalanced and sparse original datasets, 
which can be adapted to any model for any scRNA-seq data analysis task, while also achieving cross-organization privacy data resource sharing between datasets.

Due to the inherently challenging characteristics of scRNA-seq data, foundation models~\cite{yang2022scbert,cui2024scgpt,heimberg2024cell,szalata2024transformers} have gained increasing prominence recently, which have emergent knowledge beyond the original downstream dataset information. 
Specifically, scRNA-seq foundation model is a self-supervision deep learning model trained on the broad scRNA-seq dataset, with extensive semantic knowledge, that can be effectively adapted to a wide range of downstream analysis tasks.
Unlike existing dataset distillation methods, where image data is trained from scratch and pixel values are directly updated.
The challenge in scRNA-seq dataset distillation lies in distilling the foundation model knowledge and original dataset information without compromising the scRNA-seq data characteristics, which is shown in Fig~\ref{tensor}.

The detailed challenges are as follows:
(1) Directly updating gene expression values at the scRNA-seq data-level will cause them to loss inherently characteristics (\textit{e.g.,} gene expression distribution and category characteristics).
Existing dataset distillation methods are predominantly developed for image~\cite{zhao2021dataset,zhao2023dataset} and text~\cite{maekawa2023dataset,tao2024textual} focusing on distilling image pixel values and model representation spaces respectively.
Since the gene expression values in scRNA-seq data are high-dimensional, sparse and non-negative, directly distilling the original dataset information and foundational model knowledge onto scRNA-seq data-level to update them would significantly disrupt their characteristics, weakening the generalization of synthetic datasets adapted to other unseen evaluation models.
(2) Distillation with highly categorical imbalance problem leads to a loss of inter-class discriminability. \textit{i.e.,} the category characteristics of distilled synthetic datasets become increasingly blurred and tend to favor the dominant categories. 
This is due to the highly imbalanced category distribution in the original scRNA-seq dataset, where the imbalanced distillation information is back-propagated as a unified gradient across the entire synthetic dataset.

To address the above two challenges, 
(1) We first propose a latent codes based scRNA-seq dataset distillation framework named scDD, which replaces the direct update of gene expression values at the high-dimensional sparse and non-negative scRNA-seq data-level. Instead, scDD transfers and distills foundation model knowledge and original dataset information into a compact latent space and delivers the final synthetic dataset through a generator.
The synthetic dataset generated by our scDD avoids the distortion of gene expression distribution, exhibiting better scRNA-seq data characteristics and generalization to other unseen evaluation models.
(2) We then propose a single-step conditional diffusion generator named SCDG, which differs from traditional diffusion models that involve multi-step forward and reverse diffusion process.
SCDG can perform single-step gradient back-propagation, helping scDD optimize distillation quality and avoid gradient decay caused by multi-step back-propagation. Meanwhile, SCDG ensures the scRNA-seq data characteristics and inter-class discriminability of the synthetic dataset through flexible conditional control and generation quality assurance.

Our contributions are summarized as follows:
\begin{itemize}
    \item To the best of our knowledge, we are the first to consider applying dataset distillation methods to scRNA-seq data analysis tasks, and we propose a latent codes based scRNA-seq dataset distillation framework named scDD.
    \item We propose a single-step conditional diffusion generator named SCDG to enhance the category characteristics of synthetic scRNA-seq datasets, which can perform single-step gradient back-propagation to help scDD optimize the distillation quality without gradient decay.
    \item We propose a comprehensive and widely covered benchmark to evaluate the performance of scRNA-seq dataset distillation in different data analysis tasks, and the empirical results show scDD can significantly improve the distillation performance in the benchmark.
\end{itemize}

\section{Related Works}

\subsection{scRNA-seq data analysis tasks}
scRNA-seq data analysis tasks mainly include disease status classification~\cite{liu2022classification}, development stage analysis~\cite{tan2023comprehensive} and anatomical entity prediction~\cite{muijlwijk2024comparative}, among others, single-cell type annotation~\cite{cheng2023review} is a critical step in scRNA-seq data analysis. Traditional manual annotation methods typically involve two steps: first using unsupervised learning to cluster cells, and then annotating cell populations based on differentially gene expression values in each cluster.
However, this process is labor-intensive, time-consuming, and relies on prior knowledge. In contrast, automated annotation methods offer the advantages of speed and simplicity.

CellTypist~\cite{xu2023automatic} is a machine learning based automated tool for celltype annotation in scRNA-seq datasets, utilizing logistic regression classifiers optimized through the stochastic gradient descent algorithm.
scDeepInsight~\cite{jia2023scdeepinsight} is a deep learning based celltype annotation method, which first converts tabular scRNA-seq data into images through the DeepInsight methodology, enabling the use of CNNs for feature extraction and celltype classification.
SCimilarity~\cite{heimberg2024cell} is a cell atlas foundation model with encoder-decoder structure trained on a $22.7$ million cell corpus assembled across $399$ published scRNA-seq studies, which can be used to extract unifying representation and celltype annotation from gene expression profiles.
\textit{Different from them, our goal is to use foundation model to distill/condense synthetic scRNA-seq dataset for multiple data analysis tasks.}

\subsection{Dataset distillation target and pipeline}
Dataset distillation was first formally proposed by Wang~\textit{et al.}~\cite{wang2018dataset} to iteratively distill synthetic dataset using meta-learning framework~\cite{hospedales2021meta} with high computational cost. Then, dataset distillation matching target shifted toward aligning gradients DC~\cite{zhao2021dataset} and distributions DM~\cite{zhao2023dataset} between the student network and the expert network.
In recent works, the matching target such as DC and DM have been fixed, with the main focus on optimizing the dataset distillation pipeline.

\begin{figure*}[ht]
        \centering
        \subcaptionbox{Single-step conditional diffusion generator training}{
        \includegraphics[width=3.498in]{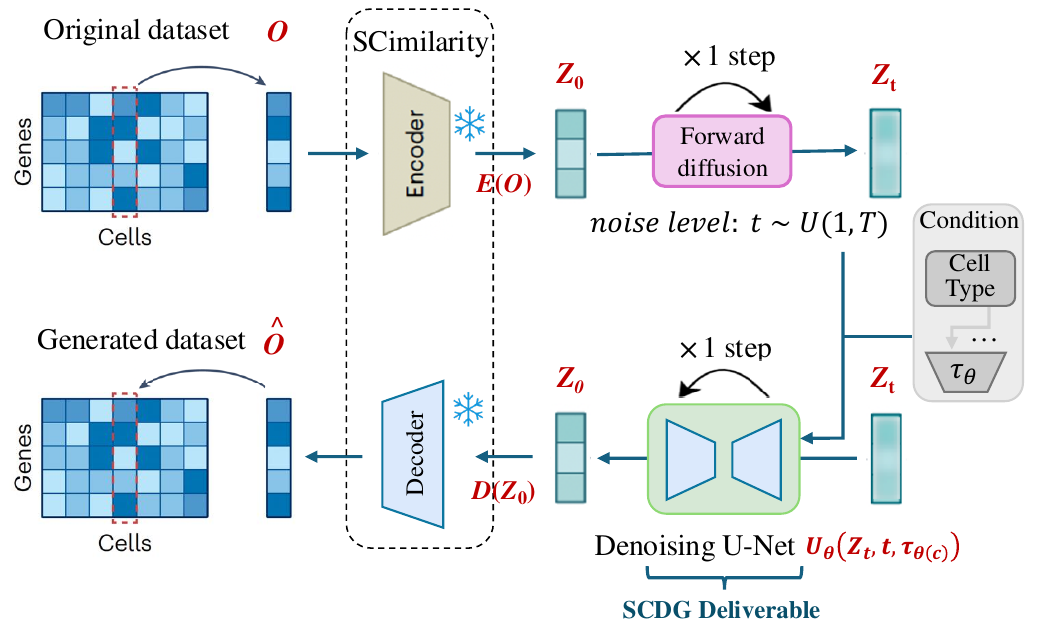}
    }\hspace{-0.48cm}
        \subcaptionbox{Latent codes based scRNA-seq dataset distillation}{
        \centering
        \includegraphics[width=3.588in]{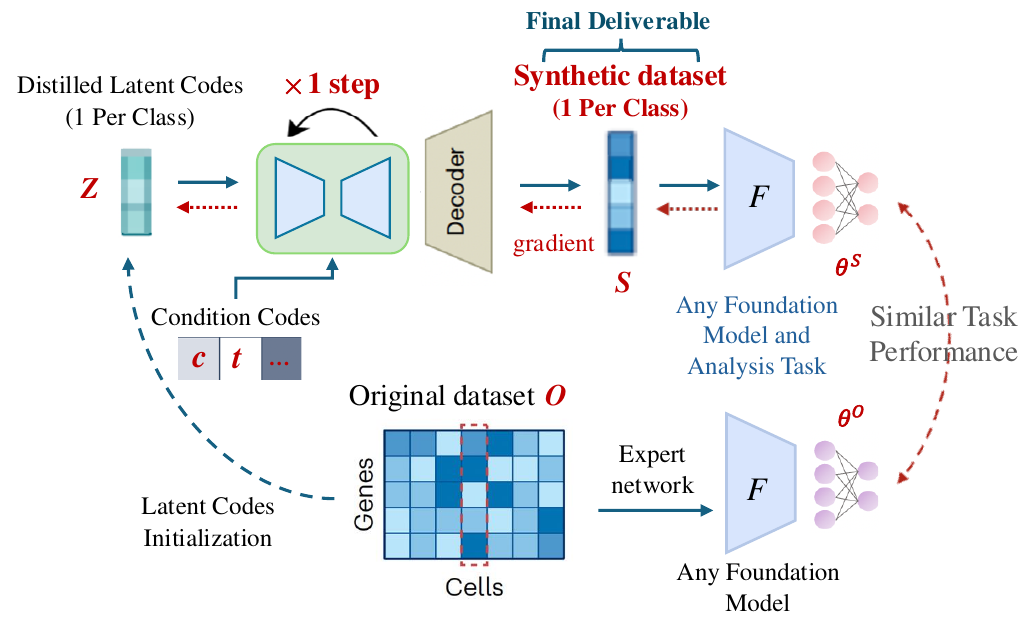}
    }
    \caption{scDD overall framework. (a) SCDG generator is delivered combining the diffusion and foundation models to generate high-quality scRNA-seq data with flexible controlled conditions, which can perform single-step gradient back-propagation to help optimize the distillation quality without gradient decay. (b) scDD finally delivers a small-scale and desensitized synthetic scRNA-seq dataset, which can replace large-scale, high-dimensional, sparse and noisy original dataset, adapt to any foundation model to achieve different scRNA-seq data analysis tasks. }
    \label{overall}
\end{figure*}

Glad~\cite{cazenavette2023generalizing} is a generative distillation pipeline that uses StyleGAN to generate synthetic images. Unlike DC and DM, which directly update the pixel values of the synthetic images, Glad updates the random Gaussian noise in the latent space before the StyleGAN. However, GAN based distillation methods often suffer from poor stability and flexibility, and the overall gradient updates to the latent space can easily blur their category characteristics.
SDXL-Turbo~\cite{su2024generative} is a generative distillation pipeline with diffusion model, which does not involve the distillation matching process, making it unable to perform distillation for synthetic datasets with few samples.
FeatDistill~\cite{maekawa2023dataset} is a feature space distillation pipeline where the distilled synthetic dataset consists of feature-level variables, and it's problem is that the usability of the distilled synthetic feature set is limited.
TextDistill~\cite{tao2024textual} is a data level distillation pipeline, which further converts the synthetic feature set from FeatDistill to a data-level synthetic dataset through generator. However, since this generator is not involved in the distillation matching process, the downstream performance of generated synthetic dataset is poor.
\textit{Different from them, we introduce a conditional diffusion generator with
single-step gradient back-propagation into the generative dataset distillation pipeline, guiding and enhancing the diversity and category characteristics of the synthetic scRNA-seq dataset in the distillation matching process.}

\section{Methods}

\subsection{scDD Problem Definition and Overall}

\subsubsection{Problem Definition}
we define the original and synthetic scRNA-seq dataset as $\mathcal{O}=\left\{\left(o_i, y_i\right)\right\}_{i=1}^{|\mathcal{O}|}$ and $\mathcal{S}=\left\{\left(s_i, y_i\right)\right\}_{i=1}^{|\mathcal{S}|}$ respectively, where the scRNA-seq data $o_{i}, s_{i} \in \mathbb{R}^d$, the category labels $y_i \in \mathcal{Y}=\{0,1, \ldots, C-1\}$, and $C$ is the number of categories, $\mathcal{O}_C$ and $\mathcal{S}_C$ are the set of original and synthetic data samples belonging to category $C$ separately.
SPC is the number of synthetic scRNA-seq data pairs for each category, \textit{i.e.,} $|\mathcal{S}|=$SPC$\times C \ll |\mathcal{O}|$. 
The goal of dataset distillation is to learn a synthetic dataset $\mathcal{S}$ that mimics the model performances trained on the original dataset $\mathcal{O}$ under the same configurations. 
Instead of directly distilling foundation model knowledge and original dataset information into the gene expression values of scRNA-seq data, 
our proposed the scDD condenses them into a compact latent space $Z$ and generates cell-level synthetic datasets $\mathcal{S}$ through a generator.

\subsubsection{Overall Framework}
our proposed the scRNA-seq dataset distillation framework scDD includes two stages: (1) single-step conditional diffusion generator training; (2) latent codes based scRNA-seq dataset distillation.
Among them, the former delivers a SCDG generator combining the diffusion and foundation models to generate high-quality scRNA-seq data with flexible controlled conditions, which can perform single-step gradient back-propagation to help optimize the distillation quality without gradient decay.
The latter finally delivers a small-scale and desensitized synthetic dataset, which can replace large-scale, high-dimensional, sparse and noisy original dataset, adapt to any foundation model to achieve any scRNA-seq data analysis tasks. The detailed scDD overall framework is shown in Fig\ref{overall}.

\subsection{Single-step conditional diffusion generator}

For scRNA-seq data generation, we follow the LDM~\cite{rombach2022high,ramesh2022hierarchical} to perform the forward and reverse diffusion processes in the latent space, training a single-step conditional diffusion generator (SCDG). Compared to GAN~\cite{goodfellow2014generative} and DDPM~\cite{ho2020denoising}, LDM offers better computational efficiency, greater robustness to high-dimensional sparse data, and more flexible conditional control~\cite{saxena2021generative,yang2023diffusion}. 
The key difference between our SCDG and LDM lies in the diffusion process: SCDG repeatedly performs a single-step forward and reverse diffusion from $1$ to $t$ and back to $1$, whereas LDM uses a continuous multi-step diffusion process from $t=\{1,2,..,T\}$ and back to $\{T,..,2,1\}$ once.
\textit{i.e.,} the forward and reverse diffusion processes in LDM are sequentially and continuously, while the two processes in our SCDG are iterative and alternating.

Meanwhile, thanks to the SCimilarity~\cite{heimberg2024cell}, we have a stable and effective encoder-decoder foundation model weight specifically for scRNA-seq data, which help us map the original scRNA-seq data $\mathcal{O}$ to the latent space $Z$.
More precisely, given a original data $\mathcal{O}$ in cell-level space, the encoder $E$ encodes it into a latent-level embedding $Z=E(\mathcal{O})$, and the decoder $D$ can reconstruct this generated scRNA-seq data $\hat{\mathcal{O}}$ from its latent embedding. \textit{i.e.,} $\hat{\mathcal{O}}=D(Z)=D(E(\mathcal{O}))$.
We perform the forward and reverse diffusion process iteratively in the $Z$ latent space .

\subsubsection{Forward Diffusion Process}
the original latent embedding $Z_0$ becomes a Gaussian noised embedding $Z_T$ by iteratively adding noise through $T$ steps. For the $t$-th step, the embedding $Z_t$ is sampled from the following distribution:
$q\left(Z_t \mid Z_{t-1}\right)=\mathcal{N}\left(Z_t ; \sqrt{1-\beta_t} Z_{t-1}, \beta_t \mathrm{I}\right),\ t=\{1,2,..,T\}.$
where $\mathrm{I}$ stands for standard Gaussian noise. $\beta_t = \beta_{\text{min}} + \frac{t-1}{T-1} \left( \beta_{\text{max}} - \beta_{\text{min}} \right)$ is a noise schedule parameter that varies with time step, $\beta_{\text {min }}$ and $\beta_{\text {max }}$ are two parameters that control the scale of $\beta_t$ in the diffusion process.
Based on the properties of the Gaussian distribution, we can bypass the step-by-step recursive computation and directly sample the noisy result $Z_t$, with the distribution given by:
\begin{equation}
q\left(Z_t \mid Z_0\right)=\mathcal{N}\left(Z_t ; \sqrt{\tilde{\alpha}_t} Z_0,\left(1-\bar{\alpha}_t\right) \mathrm{I}\right),\ t \sim \mathcal{U}(1, T).
\end{equation} 
where decay factor $\alpha_t:=1-\beta_t$ and cumulative decay factor $\bar{\alpha}_t:=\prod_{i=1}^t \alpha_i$, $Z_t$ can be directly generated by using the reparameterization~\cite{kingma2013auto}: $Z_t=\sqrt{\bar{\alpha}_t} Z_0+\sqrt{1-\bar{\alpha}_t} \epsilon$ and $\epsilon \sim \mathcal{N}(0, \mathrm{I})$.
By adjusting $\beta_t$ it can ensure that $\bar{\alpha}_T$ at the final timestep $T$ approaches zero, meaning that the endpoint $Z_T$ of the forward diffusion process approximates a standard normal distribution, \textit{i.e.,}
\begin{equation}
Z_T \approx \sqrt{\bar{\alpha}_T} Z_0+\sqrt{1-\bar{\alpha}_T} \epsilon,\  \approx \epsilon \sim \mathcal{N}(0, \mathrm{I})
\end{equation}

\subsubsection{Reverse Diffusion Process}
The training objective of the diffusion model is to learn the reverse diffusion process $p\left(Z_{t-1} \mid Z_t\right)$. Since the noise variance $\beta_t$ at each step $t$ is sufficiently small, the posterior distribution $p\left(Z_{t-1} \mid Z_t, Z_0\right)$ can be approximated as a Gaussian distribution~\cite{sohl2015deep}, making it feasible to learn the reverse process via a deep network.

In SCDG, we train a denoising deep network $\mathcal{U}_\theta$ to recover $Z_0$ from the noisy input $Z_t$, given the timestep $t$ and conditioning information $c$ (\textit{e.g.,} cell-type, organ, development stage, \textit{etc.}). Our training objective is formulated as:
\begin{equation}
\mathcal{L}_{\text {SCDG}}=\mathbb{E}_{\mathcal{U}(Z), \epsilon, t, c}\left\|Z_0-\mathcal{U}_\theta\left(Z_t, t, \tau_\theta(c) \right)\right\|^2
\end{equation}
where $Z_t$ is generated through the forward process $q(Z_t \mid Z_0)$, and $\tau_\theta(c)$ represents the learned embedding of the condition $c$. The network $\mathcal{U}_\theta$ is trained to directly predict $Z_0$, and
this mean squared error (MSE) loss $\mathcal{L}_{\text{SCDG}}$ is a simplified version of the variational lower bound. Unlike DDPM~\cite{ho2020denoising} and LDM~\cite{rombach2022high}, which reverse the diffusion process by predicting the $\epsilon-$noise instead of the clean sample, SCDG directly restores $Z_0$ and bypasses explicit noise prediction. 

Essentially, these two approaches represent different parameterizations of the same process: $\mathcal{U}_\theta\left(Z_t, t, \tau_\theta(c) \right) = Z_0 = \frac{1}{\sqrt{\alpha_t}}\left(Z_t-\sqrt{1-\alpha_t} \epsilon_\theta\left(Z_t, t, \tau_\theta(c)\right)\right)$.
\textit{i.e.,} the noise prediction network $\epsilon_\theta$ and the direct \( Z_0 \) prediction network \( \mathcal{U}_\theta \) can be transformed into each other via a linear transformation, and this formulation highlights that SCDG provides an alternative parameterization for the reverse diffusion process while maintaining theoretical equivalence with existing approaches.

\begin{algorithm}[t]
\caption{Latent codes based scRNA-seq dataset distillation framework (scDD)}
\label{algorithm}
\begin{algorithmic}[0]
\STATE \textbf{Input:} 
$\mathcal{O}$: original dataset;
$K$: total of the distillation step;
$Z^k$: latent codes in the $k$-th step;
$\mathcal{S}^k$: synthetic dataset in the $k$-th step;
$G_{\text{SCDG}}$: pre-trained single-step conditional diffusion generator;
$F$: foundation model;
$\mathcal{O}_C$: subset of $\mathcal{O}$ in category $C$;
$\mathcal{S}_C$: subset of $\mathcal{S}$ in category $C$;
$\theta$: network parameters of task header;
$\ell$: classification cross-entropy loss;
$\nabla_\theta$: gradient of $\theta$;
$\eta$: learning rate.
\STATE \textbf{Initialize:} sampling data $\mathcal{S}_0 \sim \mathcal{O}$;
\STATE \% Latent codes random initialization
\STATE Initial latent codes $Z^0=E(\mathcal{S}^0)$;
\FOR{each distillation step $k=0$ to $K$}
    \STATE \% Condition guided scRNA-seq generation
    \STATE Synthetic dataset $\mathcal{S}^k=G_{\text{SCDG}}(Z^k)$.
    \STATE \% Distribution distillation loss
    \STATE $\sum_C\left\|\frac{1}{\left|\mathcal{O}_C\right|} \sum_{o_i \in \mathcal{O}_C} F(o_i)-\frac{1}{\left|\mathcal{S}_{C}^k\right|} \sum_{s_i^k \in \mathcal{S}_C^k} F(s_i^k)\right\|$;
    \FOR{each classification epoch $n=0$ to $N$}    
        \STATE Student model: $\theta_{n+1}^{\mathcal{S}^k} \leftarrow \theta_n^{\mathcal{S}^k}-\eta_{\theta} \nabla_{\theta} \ell\left(\theta_n^{\mathcal{S}^k}\right)$;
        \STATE Expert model: $\theta_{k+1}^\mathcal{O} \leftarrow \theta_k^{\mathcal{O}}-\eta_{\theta} \nabla_{\theta} \ell\left(\theta_k^\mathcal{O}\right)$;
            \STATE \% Gradient distillation loss
    \STATE $1-\nabla_\theta \ell(\theta_k^{\mathcal{S}^k}) \cdot \nabla_\theta \ell(\theta_k^{\mathcal{O}})\ / \ {\left\|\nabla_\theta \ell(\theta_k^{\mathcal{S}^k})\right\| \left\|\nabla_\theta \ell(\theta_k^{\mathcal{O}})\right\|}$;
    \ENDFOR
    \STATE Update $Z^k$ with respect to distillation loss;
\ENDFOR
\STATE \textbf{Output:} synthetic dataset $\mathcal{S}^k=G_{\text{SCDG}}(Z^k)$; 
\end{algorithmic}
\end{algorithm}

\subsection{Latent codes based scRNA-seq distillation}

For scRNA-seq dataset distillation, instead of directly optimizing the synthetic dataset $\mathcal{S}$, we update the latent codes $Z$ in each distillation step.
To distinguish it from the diffusion process, we define $Z^0$ as the initial latent codes, $Z^k$ as the updated latent codes in the $k$-th distillation step, and the pre-trained SCDG generator $G_{\text{SCDG}}$ is used to generate the synthetic dataset $\mathcal{S}^k$. The foundation model is denoted as $F$, the network parameters of the task header is $\theta$, and its classification cross-entropy loss and the $k$-th iteration of gradient descent are $\ell$ and $\theta_k$ respectively.
$\nabla_\theta \ell(\theta^{\mathcal{S}^k}_n)$ and $\nabla_\theta \ell(\theta^{\mathcal{O}}_n)$ represent the gradients of the student model trained on $\mathcal{S}^k$ and expert model trained on $\mathcal{O}$ at the $n$-th classification epoch.
The process of scDD is divided into four parts, and the details are depicted in Algorithm~\ref{algorithm}.

\subsubsection{Latent Codes Random Initialization}

To obtain the initial latent codes $Z^0$, we first randomly sample SPC instances per class from the original dataset $\mathcal{O}$ along with their corresponding conditional information. The sampled scRNA-seq data and conditional information are then processed through the pre-trained encoder $E$ and embedding network $\tau_\theta(c)$, respectively, to generate the initial latent codes and the embedded conditional information codes.

\subsubsection{Condition Guided scRNA-seq Generation}

After obtaining the initial latent codes, we perform a total of the $K$ distillation steps. For the latent codes $Z^k$ updated in the $k$-th step, we generate the condition guided synthetic dataset $\mathcal{S}^k$ using the pre-trained SCDG generator, which is composed of the denoising deep network $\mathcal{U}$, decoder $D$ and encoder $E$, \textit{i.e., $\mathcal{S}^k=G_{\text{SCDG}}(Z^k)=D(\mathcal{U}(Z^k,t,\tau_\theta(c)))$}. Each synthetic data in $\mathcal{S}^k$ is uniquely associated with its corresponding latent code, and the final synthetic dataset $\mathcal{S}^K$ is delivered in the $K$-th step to replace the original dataset.

\subsubsection{Matching Information Distillation}

After obtaining the synthetic dataset $\mathcal{S}^k$, we match and align its gradients or distributional information with that of the original dataset $\mathcal{O}$. This matching information is then back-propagated to update the latent codes $Z^k$, ensuring that the $\mathcal{S}$ exhibits similar performance to the $\mathcal{O}$ when faced with the same task. 
Specifically, for distribution matching $\mathcal{L}_{\mathrm{DM}}$~\cite{zhao2023dataset}, 
we minimize the difference between the feature distributions of $\mathcal{S}^k$ and $\mathcal{O}$ inferred on the foundation model $F$ for each class $C$:
\begin{equation}
\mathcal{L}_{\mathrm{DM}}=\sum_C\left\|\frac{1}{\left|\mathcal{O}_C\right|} \sum_{o_i \in \mathcal{O}_C} F(o_i)-\frac{1}{\left|\mathcal{S}_{C}^k\right|} \sum_{s_i^k \in \mathcal{S}_C^k} F(s_i^k)\right\|.
\end{equation}
for gradient matching $\mathcal{L}_{\mathrm{DC}}$~\cite{zhao2021dataset}, we minimize the difference between the gradient directions of $\mathcal{S}^k$ trained on student model and $\mathcal{O}$ trained on expert model for each classification epoch:
\begin{equation}
\mathcal{L}_{\mathrm{DC}}=1-\frac{\nabla_\theta \ell(\theta_n^{{\mathcal{S}^k}}) \cdot \nabla_\theta \ell(\theta_n^{\mathcal{O}})}{\left\|\nabla_\theta \ell(\theta_n^{\mathcal{S}^k})\right\| \left\|\nabla_\theta \ell(\theta_n^{\mathcal{O}})\right\|}.
\end{equation}

\subsubsection{Distillation Information Back-propagated}

After obtaining the matching information, 
we adopt gradient checkpointing to efficiently back-propagate it for distillation with limited GPUs. To generate the synthetic dataset \(\mathcal{S}\), we do not track any gradients, compute the distillation loss $\mathcal{L}$ and its gradient with respect to the synthetic dataset, $\partial \mathcal{L} / \partial \mathcal{S}$, and then discard the computation graph used for these calculations.  
To obtain $\partial \mathcal{L} / \partial Z$, we perform a forward pass through $G_{\text{SCDG}}$, yielding $\mathcal{S} = G_{\text{SCDG}}(Z)$, while keeping track of gradients to determine $\partial \mathcal{S} / \partial Z$. By applying the chain rule, we compute:  
$\partial \mathcal{L} / \partial Z = \partial \mathcal{L} / \partial \mathcal{S} \cdot \partial \mathcal{S} / \partial Z$.  
Finally, this gradient is used to update the latent codes corresponding to synthetic data.
\section{Experiments}

\subsection{Benchmark and Datasets}

Due to the high sensitivity of scRNA-seq data to subtle variations in the experimental conditions, batch effects are often introduced by non-biological factors. \textit{e.g.,} data from different batches may exhibit different gene expressions due to slight differences in reaction conditions or data processing.
To fairly validate the effectiveness of our proposed method, we collected a comprehensive covered public datasets from different data storage management platforms (such as the Human Cell Atlas~\cite{parums2025human} and CZ CELLxGENE~\cite{czi2025cz}), and sequencing technology standards (including the 10X Gene Expression~\cite{yamawaki2021systematic} and Visium Spatial Gene Expression~\cite{rao2020bridging}). The core information of datasets is shown in Tab~\ref{datasets}.

(1) STIZ-Kidney~\cite{stewart2019spatiotemporal} contains the spatio-temporal immune topology of $27$ thousand fetal kidney cells corresponding to $44$ cell types and $4$ development stages from cortex of kidney, renal medulla, renal pelvis and ureter organ regions.

(2) SCPF-Lung~\cite{habermann2020single} is a large scRNA-seq dataset that contains $114$ thousand cells of human lungs with pulmonary fibrosis, with $27281$ gene expressions and $31$ cell types.

(3) SARS-Mouth~\cite{huang2021sars} is scRNA-seq dataset of human minor salivary glands and gingiva with $22$ thousand cell numbers and $37$ thousand gene expressions used for analyzing Severe acute respiratory syndrome coronavirus 2 (SARS-CoV-2).

(4) RSC-Cardiac\cite{simonson2023single} is the transcriptomes of over $99$ thousand human cardiac nuclei from the non-infarct region of the left ventricle of $7$ ischemic cardiomyopathy (ICM) transplant recipients and $8$ non-failing (NF) controls.

(5) SMC-Blood~\cite{arutyunyan2023spatial} is a spatially resolved multiomics single-cell atlas of the entire human maternal–fetal interface, which includes
$11$ cell types from normal and edwards syndrome populations, as well as $6$ organ anatomical entity.

(6) CTM-Brain~\cite{lerma2024cell} is the single-nucleus and spatial transcriptomics from brain subcortical multiple sclerosis and corresponding control tissues, including $62$ cell types and $10$ disease status in white matter region.

\begin{figure}[t]
        \centering
        {
        \includegraphics[width=3.533in]{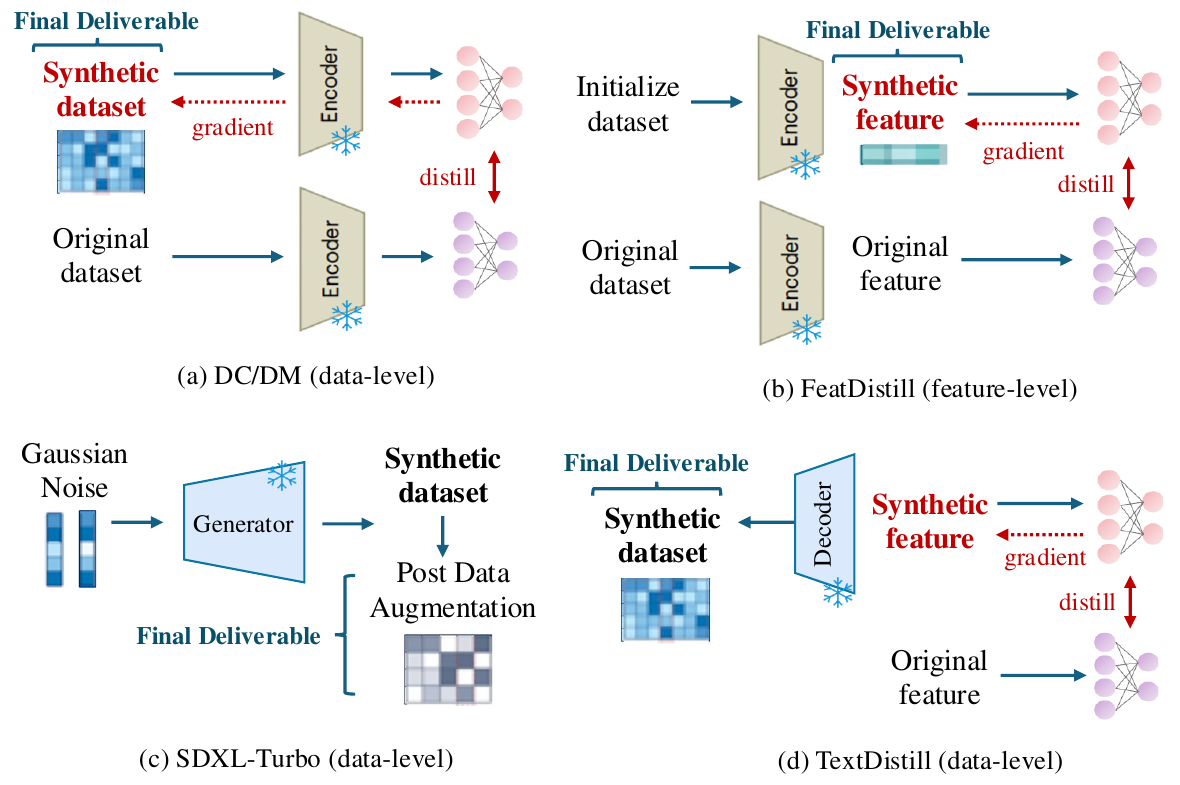}
    }
    \caption{Different distillation matching targets and optimization pipelines. (a) DC/DM distills a data-level synthetic dataset by matching gradient and distribution targets. (b) FeatDistill performs distillation optimization on feature-level and delivers synthetic features under encoder. (c) SDXL-Turbo generates synthetic dataset for data augmentation but does not involve distillation matching. (d) TextDistill distills on feature-level but delivers data-level synthetic scRNA-seq data by decoder.}
    \label{exp}
\end{figure}

\begin{table*}[h]
\footnotesize
\centering
\caption{Statistics of the scRNA-seq datasets used for the proposed benchmark with comprehensive coverage.}
\label{datasets}
\renewcommand{\arraystretch}{1.2} % Increase spacing between horizontal lines
\resizebox{0.888\linewidth}{!}{
\begin{tabular}{c|c|c|c|c|c}
\hline
\textbf{Dataset Name} & \textbf{Tissue Region}  & \textbf{Disease}  & \textbf{Celltype Number} & \textbf{Sample Number} & \textbf{Gene Expression} \\
\hline
STIZ-Kidney~\cite{stewart2019spatiotemporal} & Kidney & Normal & Multi-Class (44)  & $27,203$  & $33,694$ \\
SCPF-Lung~\cite{habermann2020single} & Lung & Fibrosis & Multi-Class (31)  & $114,396$  & $27,281$ \\
SARS-Mouth~\cite{huang2021sars} & Mouth & Covid & Multi-Class (62)  & $22,472$  & $37,731$ \\
RSC-Cardiac\cite{simonson2023single} & Cardiac  & Ischemia & Multi-Class (16)  & $99,684$  & $36,601$ \\
SMC-Blood~\cite{arutyunyan2023spatial} & Blood  & Trisomy & Multi-Class (11)  & $64,070$  & $33,538$ \\
CTM-Brain~\cite{lerma2024cell} & Brain & Sclerosis & Multi-Class (62)  & $95,334$  & $30,242$ \\
\hline
\end{tabular}}
\end{table*}

\begin{table*}[h]
\centering
\caption{Performances of single-cell type annotation using various dataset distillation methods with SPC=1 on the proposed benchmark.}
\label{result_1}
\renewcommand{\arraystretch}{1.4} 
\resizebox{0.98\linewidth}{!}{
\begin{tabular}{lc|c|c|c|c|c|c}
\hline \multicolumn{2}{c}{\textbf{DATASET}} & \multicolumn{1}{|c}{ \textbf{STIZ-Kidney}} & \multicolumn{1}{|c}{ \textbf{SCPF-Lung} } & \multicolumn{1}{|c}{ \textbf{SARS-Mouth}}  & \multicolumn{1}{|c}{\textbf{RSC-Cardiac}} & \multicolumn{1}{|c}{\textbf{ SMC-Blood}} & \multicolumn{1}{|c}{\textbf{CTM-Brain }} \\

\multicolumn{2}{c|}{SPC (Condense ratio)} & 1\ ($1.62$ \text{‰}) & 1\ ($0.27$ \text{‰})  & 1\ ($2.76$ \text{‰})  & 1\ ($0.16$ \text{‰}) & 1\ ($0.17$ \text{‰}) & 1\ ($0.65$ \text{‰}) \\ 
\hline \hline 
 
\multicolumn{2}{c|}{CellTypist~\cite{xu2023automatic} }& 
$20.35_{\pm 1.74}$  & 
$39.81_{\pm 5.86}$ & 
$25.11_{\pm 2.87}$ &  
$26.08 _{\pm 7.07}$    & 
$37.78_{\pm 2.39}$ &  
$19.54 _{\pm 2.22}$  \\
 
\multicolumn{2}{c|}{scDeepInsight~\cite{jia2023scdeepinsight} }& 
$24.93_{\pm 0.01}$ & 
$37.40_{\pm 0.01}$  &
$27.05_{\pm 0.01}$ & 
$55.03 _{\pm 0.01}$ & 
$32.63 _{\pm 0.01}$ & 
$17.42 _{\pm 0.01}$\\ 
\multicolumn{2}{c|}{SCimilarity~\cite{heimberg2024cell} }& $54.77_{\pm 0.01}$ & $50.11_{\pm 0.01}$  & $38.13_{\pm 0.01}$ & $78.20 _{\pm 0.01}$ & $34.99 _{\pm 0.01}$ & $32.72 _{\pm 0.03}$ \\
\hline

\multicolumn{2}{c|}{SDXL-Turbo~\cite{su2024generative} }& $62.03 _{\pm 0.01}$ & $52.71_{\pm 0.01}$  & $44.37_{\pm 0.01}$ & $84.02 _{\pm 0.01}$ & $31.69 _{\pm 0.01}$ & $32.11 _{\pm 0.02}$\\

\multicolumn{2}{c|}{DC~\cite{zhao2021dataset} }& $72.51 _{\pm 0.01}$ & $74.54_{\pm 0.01}$  & $55.02_{\pm 0.01}$ & $86.75 _{\pm 0.01}$ & $42.36 _{\pm 0.04}$ & $43.89 _{\pm 0.01}$ \\
 
\multicolumn{2}{c|}{DM~\cite{zhao2023dataset} }& $73.51 _{\pm 0.01}$ & $83.46_{\pm 0.01}$  & $56.88_{\pm 0.01}$ & $90.27 _{\pm 0.01}$ & $61.40 _{\pm 0.01}$ & $50.69_{\pm 0.01}$ \\

\hline
\multicolumn{2}{c|}{FeatDistill~\cite{maekawa2023dataset} }
& $71.56 _{\pm 0.01}$  & $80.17_{\pm 0.01}$ & $49.63_{\pm 0.01}$ & $86.52_{\pm 0.01}$ & $55.60 _{\pm 0.01}$ & $40.14_{\pm 0.01}$\\\

+ DC  & + DM   &  $71.18_{\pm 0.01}$ & $84.81_{\pm 0.01}$  & $57.15_{\pm 0.01}$ &  $87.31 _{\pm 0.01}$ &  $61.10_{\pm 0.01}$ &  $51.17_{\pm 0.01}$\\

%\hline
\multicolumn{2}{c|}{TextDistill~\cite{tao2024textual} } & $64.65_{\pm 0.01}$ & $74.87_{\pm 0.01}$ & $52.97_{\pm 0.01}$ & $82.91_{\pm 0.01}$  & $35.41_{\pm 0.01}$ & $36.19_{\pm 0.01}$\\
 
\ + DC  &  + DM  &  $57.31_{\pm 0.01}$ & $71.53_{\pm 0.01}$ &    $46.46_{\pm 0.01}$ &  $83.19_{\pm 0.01}$  & $39.69_{\pm 0.01}$ &  $33.82 _{\pm 0.01}$\\

\hline \hline 

\multicolumn{2}{c|}{\textbf{scDD} } & $\textbf{75.61}_{ \pm \textbf{ 0.01}}$ & $\textbf{87.39} _{\pm \textbf{0.01}}$  &  $\textbf{61.15}_{\pm \textbf{0.01}}$ & $\textbf{89.81} _{\pm\textbf{0.01}}$ & $\textbf{61.28} _{\pm\textbf{ 0.01}}$ & $\textbf{52.15} _{\pm \textbf{0.01}}$\\

\hline  %\hline 
\multicolumn{2}{c|}{Original Dataset}&  $90.20 _{\pm 0.01}$ & $\ 95.05 _{\pm 0.01}$ & $\ 68.60 _{\pm 0.01}$ & $\ 97.02 _{\pm 0.01}$ & $\ 87.68 _{\pm 0.01}$ & $\ 71.52 _{\pm 0.01}$   \\
\hline  
\end{tabular}
}
\end{table*}

\subsection{Baselines and Models}

We compare our method with a series of latest and representative cell type annatation and dataset distillation methods.
(1) Single-cell type annotation method:
CellTypist~\cite{xu2023automatic} machine learning with logistic regression; scDeepInsight~\cite{jia2023scdeepinsight} deep learning with convolutional neural networks; SCimilarity~\cite{heimberg2024cell} foundation model with cell expression profiles.
(2) Dataset distillation matching target: 
DC~\cite{zhao2021dataset} gradient matching;
DM~\cite{zhao2023dataset} distribution matching;
SDXL-Turbo~\cite{su2024generative} generation mismatching.
(3) Dataset distillation optimization pipeline: 
%Glad~\cite{cazenavette2023generalizing} latent code optimization;
FeatDistill~\cite{maekawa2023dataset} feature space optimization;
TextDistill~\cite{tao2024textual} data level optimization.
The different dataset distillation methods are shown in detail in Fig~\ref{exp}.
Additionally, we also evaluate the cross-architecture generalization performance of the synthetic scRNA-seq dataset distilled by SCimilarity~\cite{heimberg2024cell} (foundation model) to the different evaluation models: including CellTypist~\cite{xu2023automatic} (machine learning), scDeepInsight~\cite{jia2023scdeepinsight} (convolutional network) and TOSICA~\cite{chen2023transformer} (transformer network).

\subsection{Implementation Details}

For all scRNA-seq datasets, we use Scanpy~\cite{wolf2018scanpy} and AnnData~\cite{virshup2021anndata} in conversion of sparsity and unstructured data to uniform .h5ad HDF5 files, and filter out cells with $< 10$ expression counts and genes that expressed in $< 3$ cells.
The datasets were split into training and testing sets with a ratio of $0.7$:$0.3$.
In the pre-processing, the order of gene expression in the scRNA-seq dataset is registered with SCimilarity, the value is normalized by $1e4$ total counts and then logarithmized.
For celltype annotation task, we use the frozen SCimilarity’s encoder to extract $128$ dimensional embedding features and add a layer of classification header for prediction.
To ensure distillation fairness in comparison, we follow up the experimental setup as stated in~\cite{cui2022dc}, in the training stage, we set the diffusion step $t$ to $1000$, $\beta_{\min }=10^{-4}$ and $\beta_{\max}=0.1$. In the evaluation stage after the synthetic dataset is generated, we train $10$ randomly initialized network on it each time for $1000$ training epochs using SGD optimizer, and report the mean and standard deviation of their accuracy on the original testset.
Notice that all experiments were run on the server with four NVIDIA A40 (48G) GPUs.

\subsection{Results}

We present the dataset distillation performance for different single-cell analysis tasks. To evaluate distillation performance, we first generate synthetic datasets through candidate methods, and train target networks using these synthetic datasets. Then, the performance of these target networks is evaluated on the corresponding original test dataset. 

\subsubsection{Single-cell type annotation}
The distillation performance results for cell-type annotation task are shown in Tab~\ref{result_1}, we condense the number of original scRNA-seq datasets on average to $0.94$ \text{‰} with SPC=1. 
It is seen that our proposed method outperforms all state-of-the-art baseline methods in each scRNA-seq dataset with a significant improvement. \textit{e.g.,} we improve the absolute performance in SCPF-Lung more than $12.07\%$, and in SMC-Blood more than $16.55\%$ on average. Meanwhile, compared to the single-cell type annotation, distillation matching target and distillation optimization pipeline methods, our proposed method achieve $34.61\%$, $9.85\%$ and $9.36\%$ absolute improvement separately on average.

\subsubsection{Other scRNA-seq data analysis tasks}
The distillation performance results for disease status classification, development stage analysis and anatomical entity prediction tasks are shown in Tab~\ref{result_2}. 
We condense the number of original scRNA-seq datasets on average to $0.35$ \text{‰} with SPC=1, 5.
Overall, our proposed method can achieve $5.64\%$ absolute and $15.81\%$ relative improvement over previous state-of-the-art methods on average tasks. Specifically, when SPC is 1, we achieve an average $4.87\%$ absolute improvement and $15.17\%$ relative improvement, and when SPC is 5, we can achieve $6.46\%$ absolute improvement and $16.02\%$ relative improvement on average. Meanwhile, increasing the condense ratio of the synthetic scRNA-seq dataset from $0.11$ \text{‰} to $0.58$ \text{‰} will result in an average $9.49\%$ absolute improvement and $25.01\%$ relative improvement respectively.

\begin{table*}[t]
\centering
\caption{Performances of other scRNA-seq data analysis tasks using various dataset distillation methods with SPC=1, 5 on the proposed benchmark.}
\label{result_2}
\renewcommand{\arraystretch}{1.48} 
\resizebox{1.0\linewidth}{!}{
\begin{tabular}{lc|c|c|c|c|c|c|c}

\hline \multicolumn{2}{c}{\textbf{OTHER TASKS}} & \multicolumn{2}{|c}{ \textbf{Disease Status}}  & \multicolumn{2}{|c}{ \textbf{Development Stage}}   & \multicolumn{2}{|c}{\textbf{ Anatomical Entity}}  &  \\
\multicolumn{2}{c}{DATASET (Class)} & \multicolumn{2}{|c}{ CTM-Brain\ (10)}  & \multicolumn{2}{|c}{ STIZ-Kidney\ (4)}   & \multicolumn{2}{|c|}{SMC-Blood\ (6)}  & Average \\
\multicolumn{2}{c}{SPC (Condense ratio)} & 
\multicolumn{2}{|c}{ 1\ ($0.10$  \text{‰}) \ \    5\ ($0.52$ \text{‰}) }
  &  
\multicolumn{2}{|c}{ 1\ ($0.15$ \text{‰}) \ \    5\ ($0.74$ \text{‰}) }
 &  
 \multicolumn{2}{|c|}{ 1\ ($0.09$ \text{‰}) \ \   5 \ ($0.47$ \text{‰}) }  &  Results
\\  \hline \hline 
\multicolumn{2}{c|}{SCimilarity~\cite{heimberg2024cell} }&
\multicolumn{2}{c|}{$12.02_{\pm 0.01}$ \ \   $23.06_{\pm 0.01}$} & 
\multicolumn{2}{c|}{ $27.62_{\pm 0.01}$  \ \   $45.71 _{\pm 0.01}$ }
 & \multicolumn{2}{c|}{ $27.36 _{\pm 0.01}$  \  \  $28.71_{\pm 0.01}$ }
 &  $27.41_{\pm 0.01}$ \\

\multicolumn{2}{c|}{DC~\cite{zhao2021dataset} }& 
\multicolumn{2}{c|}{$ 21.94_{\pm 0.01}$ \ \    $36.18_{\pm 0.01}$}& 
\multicolumn{2}{c|}{$ 43.59_{\pm 0.01}$  \ \   $56.45_{\pm 0.01}$}& 
\multicolumn{2}{c|}{$ 41.53_{\pm 0.01}$ \ \   $45.81_{\pm 0.01}$} 
 &  $40.92_{\pm 0.01}$ \\
 
\multicolumn{2}{c|}{DM~\cite{zhao2023dataset} } &
\multicolumn{2}{c|}{ $19.80_{\pm 0.01}$ \ \   $31.08_{\pm 0.01}$}& 
\multicolumn{2}{c|}{ $42.65_{\pm 0.01}$  \ \   $52.69_{\pm 0.01}$}& 
\multicolumn{2}{c|}{ $40.79 _{\pm 0.01}$  \ \    $44.54_{\pm 0.01}$} 
 &  $38.59_{\pm 0.01}$ \\

\multicolumn{2}{c|}{FeatDistill~\cite{maekawa2023dataset} }&
\multicolumn{2}{c|}{ $19.09_{\pm 0.01}$ \ \  $30.24_{\pm 0.01}$}& 
\multicolumn{2}{c|}{ $43.79 _{\pm 0.01}$  \ \   $51.55 _{\pm 0.01}$}& 
\multicolumn{2}{c|}{ $38.77_{\pm 0.01}$ \ \    $42.34 _{\pm 0.01}$} &  $37.63_{\pm 0.01}$ \\

+ DC  &   + DM     &  
\multicolumn{2}{c|}{ $19.18 _{\pm 0.01}$  \ \   $30.47_{\pm 0.01}$}& 
\multicolumn{2}{c|}{ $44.06_{\pm 0.01}$ \ \   $50.93_{\pm 0.01}$}& 
\multicolumn{2}{c|}{ $39.23_{\pm 0.01}$  \ \   $41.09_{\pm 0.01}$}&  $37.51_{\pm 0.01}$ \\

\multicolumn{2}{c|}{TextDistill~\cite{tao2024textual} } & 
\multicolumn{2}{c|}{$ 21.04_{\pm 0.01}$ \ \   $29.77_{\pm 0.01}$}& 
\multicolumn{2}{c|}{ $38.89 _{\pm 0.01}$  \ \  $49.33_{\pm 0.01}$}& 
\multicolumn{2}{c|}{ $35.61_{\pm 0.01}$ \ \   $38.81 _{\pm 0.01}$}  &  $35.58_{\pm 0.01}$ \\
 
+ DC  &  + DM     & 
\multicolumn{2}{c|}{  $21.63 _{\pm 0.01}$  \ \   $29.55_{\pm 0.01}$} & 
\multicolumn{2}{c|}{ $40.96 _{\pm 0.01}$ \ \   $50.58_{\pm 0.01}$}& 
\multicolumn{2}{c|}{  $35.24 _{\pm 0.01}$  \ \   $37.84_{\pm 0.01}$} &
$35.97_{\pm 0.01}$ \\ \hline  \hline

\multicolumn{2}{c|}{\textbf{scDD} } &  
\multicolumn{2}{c|}{ $\textbf{22.57}_{\pm \textbf{0.01}}$ \ \ \  $\textbf{36.84}_{\pm \textbf{0.01}}$}&  
\multicolumn{2}{c|}{$ \textbf{48.13}_{\pm \textbf{0.01}}$  \ \ \ $\textbf{58.27}_{\pm \textbf{0.01}}$}&   
\multicolumn{2}{c|}{$\textbf{43.08} _{\pm \textbf{0.01}}$  \ \  \
$\textbf{47.13} _{\pm \textbf{0.01}}$}&  
$\textbf{42.67} _{\pm \textbf{0.01}}$ \\
 
\hline
\multicolumn{2}{c|}{Original Dataset}& 
\multicolumn{2}{c|}{$67.79 _{\pm 0.01}$ }  &    
\multicolumn{2}{c|}{$83.63_{\pm 0.01}$ }  & 
\multicolumn{2}{c|}{$71.72 _{\pm 0.01}$ }  &
$74.38_{\pm 0.01}$ \\ \hline  

\end{tabular}
}
\end{table*}

\begin{table*}[t]
\centering
\small 
\caption{Quantitative analysis on the cross-architecture generalization of the synthetic scRNA-seq dataset distilled by SCimilarity to the various unseen evaluation models with different dataset distillation methods.}
\label{table1}
\renewcommand{\arraystretch}{1.432}
\resizebox{1.0\linewidth}{!}{
\begin{tabular}{c|cccc|cccc|cccc}
\hline & \multicolumn{12}{|c}{ \textbf{Cross-Architecture Generalization of  
the Synthetic scRNA-seq Dataset}}   \\
Evaluation model  & \multicolumn{4}{|c}{CellTypist~\cite{xu2023automatic} (Machine Learning)}  & \multicolumn{4}{|c}{scDeepInsight~\cite{jia2023scdeepinsight} (Convolutional Network)}     & \multicolumn{4}{|c}{TOSICA~\cite{chen2023transformer} (Transformer Network)}    \\
Distillation method & DC & DM  & TextDistill &  \textbf{scDD} (Avg)   & DC & DM  & TextDistill &  \textbf{scDD} (Avg)  & DC & DM  & TextDistill &  \textbf{scDD} (Avg) \\ \hline
STIZ-Kidney 
& $28.37$ & $30.63$ & $57.90$  & $\textbf{75.62}\ (\textcolor{red}{\uparrow 36.7} ) $  
& $23.47$ & $23.61$ & $59.40$  & $\textbf{70.80}\ (\textcolor{red}{\uparrow 35.3} ) $   
& $33.26$ & $34.22$ & $66.63$  & $\textbf{77.50}\ (\textcolor{red}{\uparrow 32.8} ) $  \\
SCPF-Lung 
& $06.93$ & $08.73$ & $76.27$  & $\textbf{78.99}\ (\textcolor{red}{\uparrow 48.3} ) $  
& $16.07$ & $17.56$ & $73.89$  & $\textbf{75.61}\ (\textcolor{red}{\uparrow 39.8} ) $   
& $36.25$ & $37.31$ & $80.52$  & $\textbf{79.59}\ (\textcolor{red}{\uparrow 28.2} ) $  \\
SARS-Mouth 
& $27.50$ & $30.45$ & $39.24$  & $\textbf{49.36}\ (\textcolor{red}{\uparrow 17.0} ) $  
& $20.45$ & $21.43$ & $32.30$  & $\textbf{37.54}\ (\textcolor{red}{\uparrow 12.8} ) $   
& $37.34$ & $36.83$ & $31.35$  & $\textbf{45.02}\ (\textcolor{red}{\uparrow 9.85} ) $  \\
RSC-Cardiac 
& $45.24$ & $47.24$ & $89.54$  & $\textbf{94.02}\ (\textcolor{red}{\uparrow 33.7} ) $  
& $18.56$ & $23.75$ & $80.40$  & $\textbf{91.25}\ (\textcolor{red}{\uparrow 50.3} ) $   
& $48.76$ & $54.59$ & $91.73$  & $\textbf{94.45}\ (\textcolor{red}{\uparrow 29.4} ) $  \\
SMC-Blood 
& $25.49$ & $29.87$ & $40.30$  & $\textbf{44.77}\ (\textcolor{red}{\uparrow 12.9} ) $  
& $35.29$ & $32.31$ & $23.42$  & $\textbf{45.65}\ (\textcolor{red}{\uparrow 15.3} ) $   
& $36.22$ & $32.85$ & $38.95$  & $\textbf{43.83}\ (\textcolor{red}{\uparrow 7.82} ) $  \\
CTM-Brain 
& $15.57$ & $17.66$ & $22.14$  & $\textbf{42.88}\ (\textcolor{red}{\uparrow 24.4} ) $  
& $22.01$ & $25.37$ & $32.28$  & $\textbf{44.29}\ (\textcolor{red}{\uparrow 17.7} ) $   
& $19.07$ & $17.25$ & $28.76$  & $\textbf{46.51}\ (\textcolor{red}{\uparrow 24.8} ) $  \\
\hline
\end{tabular}
}
\end{table*}

\begin{table*}[t]
\centering
\small 
\caption{Quantitative analysis on the single-step conditional diffusion generator (SCDG) for the distillation performance improvement of our scDD framework with gradient and distribution matching in the scRNA-seq data analysis tasks.}
\label{generator}
\renewcommand{\arraystretch}{1.46}
\resizebox{1.0\linewidth}{!}{
\begin{tabular}{c|ccccccccc}
\hline & \multicolumn{6}{|c}{ \textbf{Single-cell Type Annotation}} & \textbf{Disease Status} &\textbf{ Development Stage }& \textbf{ Anatomical Entity}  \\
Commponent & STIZ-Kidney  & SCPF-Lung & SARS-Mouth  & RSC-Cardiac  & SMC-Blood   & CTM-Brain & CTM-Brain  & STIZ-Kidney  & SMC-Blood  \\
\hline

DC+ & $72.51  $ & $74.54  $ & $ 55.02  $ & $86.75 $ & $42.36 $ & $43.89 $ & $21.94 $ & $43.59 $ & $41.53  $ \\

scDD+Decoder & $71.46\ (\textcolor{black}{\downarrow 1.05})  $ 
& $86.27\ (\textcolor{black}{\uparrow 11.73} )   $ 
& $61.57\ (\textcolor{black}{\uparrow 6.55} )  $ 
& $88.43\ (\textcolor{black}{\uparrow 1.68} )   $ 
& $63.56\ (\textcolor{black}{\uparrow 21.2} )   $ 
& $49.71\ (\textcolor{black}{\uparrow 5.82} )  $ 
& $20.38\ (\textcolor{black}{\downarrow 1.56})   $ 
& $46.32\ (\textcolor{black}{\uparrow 2.73})    $ 
& $42.14\ (\textcolor{black}{\uparrow 0.61})    $ \\

\textbf{scDD+SCDG} & 
$73.82\ (\textcolor{red}{\uparrow 2.36} )  $ & 
$\textbf{87.39}\ (\textcolor{red}{\uparrow 1.12}) $ & 
$\textbf{61.15}\ (\textcolor{blue}{\downarrow 0.42})   $ & 
$89.29\ (\textcolor{red}{\uparrow 0.86})  $ & 
$\textbf{65.32}\ (\textcolor{red}{\uparrow 1.76})  $ & 
$50.58\ (\textcolor{red}{\uparrow 0.87})  $ & 
$\textbf{22.57}\ (\textcolor{red}{\uparrow 2.19})  $ & 
$\textbf{48.13}\ (\textcolor{red}{\uparrow 1.81}) $ & 
$\textbf{43.08}\ (\textcolor{red}{\uparrow 0.92})  $ \\

DM+ & $73.51  $ & $83.46  $ & $ 56.88  $ & $90.27 $ & $61.40 $ & $50.69 $ & $19.80 $ & $42.65 $ & $40.79  $ \\

scDD+Decoder & 
$74.14\ (\textcolor{black}{\uparrow 0.63} )    $ & 
$84.29\ (\textcolor{black}{\uparrow 0.83} )      $ & 
$ 57.11\ (\textcolor{black}{\uparrow 0.23} )     $ & 
$89.12\ (\textcolor{black}{\downarrow 1.15} )      $ & 
$60.86\ (\textcolor{black}{\downarrow 0.54} )    $ & 
$51.69\ (\textcolor{black}{\uparrow 1.00} )      $ & 
$20.75\ (\textcolor{black}{\uparrow 0.95} )      $ & 
$43.30\ (\textcolor{black}{\uparrow 0.65} )       $ & 
$41.68\ (\textcolor{black}{\uparrow 0.89} )       $ \\

\textbf{scDD+SCDG} & 
$\textbf{75.61}\ (\textcolor{red}{\uparrow 1.47} )  $  & 
$85.92\ (\textcolor{red}{\uparrow 1.63} )   $ & 
$57.68 \ (\textcolor{red}{\uparrow 0.57} )  $ & 
$\textbf{89.81} \ (\textcolor{red}{\uparrow 0.69} )  $ & 
$61.28 \ (\textcolor{red}{\uparrow 0.42} ) $ & 
$\textbf{52.15} \ (\textcolor{red}{\uparrow 0.46} )  $  & 
$21.89 \ (\textcolor{red}{\uparrow 1.14} )  $ & 
$43.62 \ (\textcolor{red}{\uparrow 0.32} )$ & 
$42.73 \ (\textcolor{red}{\uparrow 1.05} ) $ \\

\hline
\end{tabular}
}
\end{table*}

From the distillation results and trends, we have summarized the following observations:
(1) The distillation performance based on gradient matching generally outperforms distribution matching on average, which is because distribution matching methods completely ignore the optimization process of model training and imprecise estimation for the data distribution leads to inferior result.
(2) The quality of the initial synthetic dataset significantly impacts the performance improvement from distillation, better classification performance of the initial dataset leads to greater performance gains through distillation, while lower classification performance makes distillation less effective.
(3) Our method achieves optimal performance across various scRNA-seq datasets and analysis tasks. This is partly due to our dataset distillation framework scDD, which does not directly update the gene expression values but instead indirectly updates the latent codes before the generator, and directly updating the gene expressions can disrupt their category and non-negativity characteristics, leading to suboptimal results for the analysis tasks. Moreover, our single-step conditional diffusion generator SCDG helps enhance the diversity of synthetic data, preventing inter-class synthetic data from becoming increasingly similar and loosing their category characteristics during distillation process.

\section{Analysis}

\subsection{Main Analysis}

\subsubsection{Generalization of synthetic scRNA-seq dataset} 
for dataset distillation methods that need to be deployed in practical application scenarios, a satisfactory distilled synthetic dataset should  have similar training effects to the original one on downstream models with arbitrary architectures.
To validate the cross-architecture generalization of the synthetic dataset distilled by our scDD, we train the distilled synthetic dataset with SCimilarity~\cite{heimberg2024cell} foundation model knowledge and network architecture on the unseen models~\cite{xu2023automatic,jia2023scdeepinsight,chen2023transformer} and evaluate them on the original test dataset.
The results are shown in Tab~\ref{table1}, it is obvious that compared to existing distillation methods, our proposed scDD still maintains the best performance with an average $26.51\%$ absolute improvement when faced with different unseen network architectures.
This is largely attributed to the synthetic dataset generated by the latent codes has better generalization compared to directly updating gene expression values at the data-level.

\subsubsection{Single-step conditional diffusion generator (SCDG)}
to validate the performance improvement of our proposed SCDG with gradient and distribution matching in distillation process, we compare SCDG and SCimilarity's decoder (similar to Glad~\cite{cazenavette2023generalizing} without StyleGAN) by inserting them into our scDD distillation framework as generators separately. The results are shown in Tab~\ref{generator}, it is obvious that our SCDG is better suited to the latent codes based scRNA-seq dataset distillation framework, leading to an average $1.07\%$ absolute performance improvement across various data analysis tasks.
This is due to the continuously gradient-updated latent codes, which through our SCDG generator, can produce more stable and flexible synthetic scRNA-seq data with diverse and distinct category features. In addition, this synthetic data is better suited for the different unseen evaluation models. 

\begin{table}[t]
\centering
\small 
\caption{Ablation study on the different numbers of synthetic scRNA-seq per class (SPC) for the distillation performance of our scDD in single-cell type annotation task.}
\label{spc}
\renewcommand{\arraystretch}{1.18}
\resizebox{0.96\linewidth}{!}{
\begin{tabular}{c|ccccc}
\hline & \multicolumn{5}{|c}{\textbf{Number of scRNA-seq per class}}   \\
SPC & 1  & 2 & 5  & 10 & ALL  \\
\hline
STIZ-Kidney & $75.61  $ & $77.32  $ & $81.65  $ & $83.29 $  & $90.20 $  \\
SCPF-Lung & $87.39  $ & $88.04  $ & $89.85  $ & $90.73 $ & $95.05 $\\
SARS-Mouth  & $61.15  $ & $61.59  $ & $ 63.18  $ & $64.45 $  & $68.60 $\\
RSC-Cardiac & $89.81  $ & $89.86  $ & $ 90.56  $ & $91.34 $ & $97.02 $\\
SMC-Blood & $65.32 $ & $66.44  $ & $67.91  $ & $71.77 $ & $87.68 $ \\
CTM-Brain & $52.15 $ & $53.81  $ & $ 58.53  $ & $60.63 $ & $71.52 $ \\

\hline
\end{tabular}
}
\end{table}

\subsection{Ablation Study}

\subsubsection{Number of synthetic scRNA-seq per class (SPC)}
we evaluate the effectiveness of our proposed scDD in different number of 
synthetic scRNA-seq data pairs per class, the results are shown in Tab~\ref{spc}. 
It can be observed that our method can consistently improve distillation performance with increasing SPC, and when SPC is 1 (compression ratio is $0.94 \text{‰}$), the accuracy of the distilled synthetic dataset reaches an average $84.58\%$ of the original full dataset.
As SPC increases to 2, 5, and 10, the accuracy proportion relative to the original dataset can reach $85.69\%$, $88.55\%$, and $90.62\%$ respectively. 
These results demonstrate that our method can distill a synthetic dataset with significant performance at very low compression ratios, while the distillation performance can be further improved by increasing SPC, at the cost of higher computational and memory demands.

\subsubsection{Impact of freezing the foundation model weights during distillation}
we study the performance impact of freezing the foundation model parameter weights during distillation process, the results are shown in Tab~\ref{frozen}. 
It is evident that when the foundation model weights are not frozen during the distillation process, the distillation matching loss and accuracy exhibit a significant decline compared to the frozen setting. Specifically, the loss increases by an average of $3110$ times, while the accuracy decreases by $26.85\%$. This result indicates that updating the foundation model during distillation can cause inaccurate matching for gradients or distributions, leading to a substantial performance degradation.

\subsubsection{Impact of task head's parameter size on the foundation model}
\begin{figure}[t]
        \centering
        \subcaptionbox{Distillation Accuracy}{
        \includegraphics[width=1.662in]{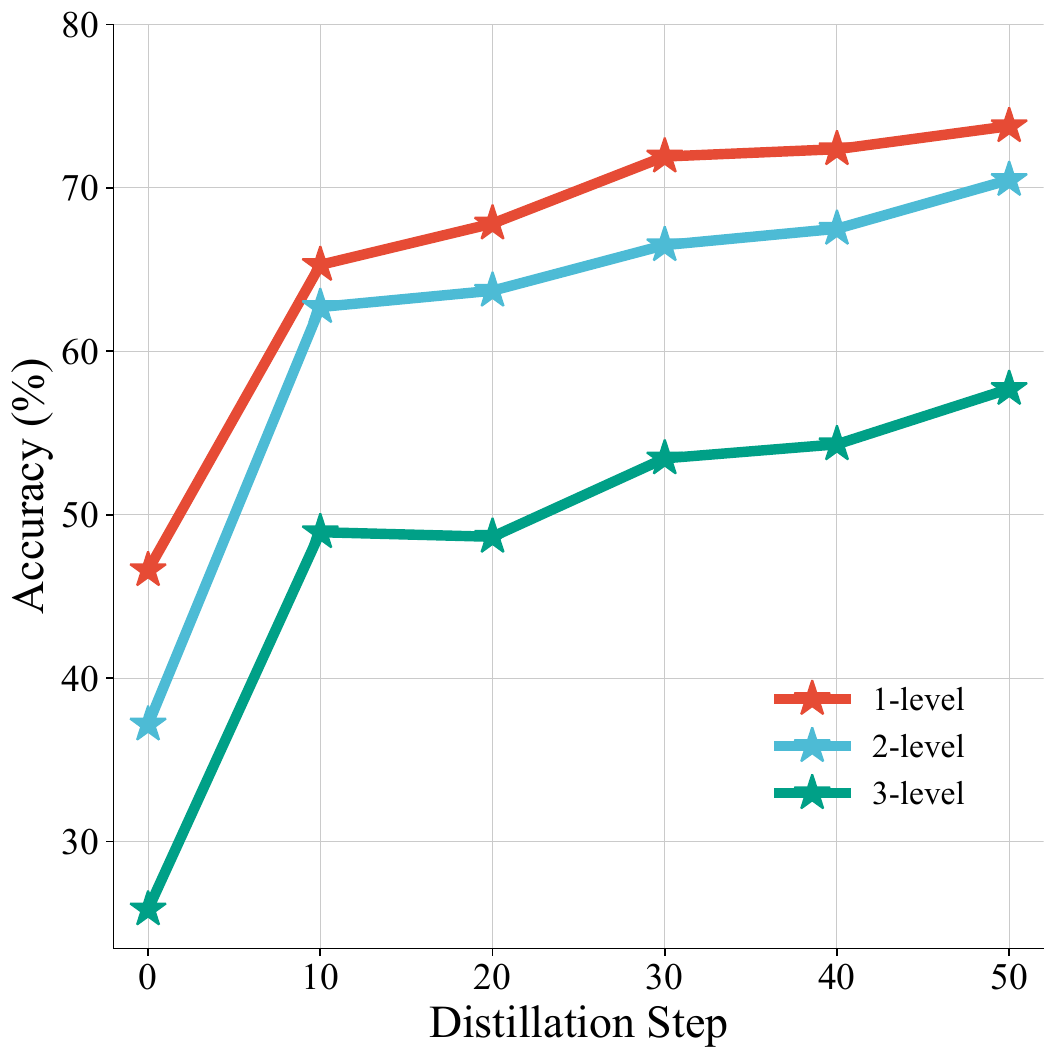}
    }\hspace{-0.19cm}
        \subcaptionbox{Matching Loss}{
        \centering
        \includegraphics[width=1.662in]{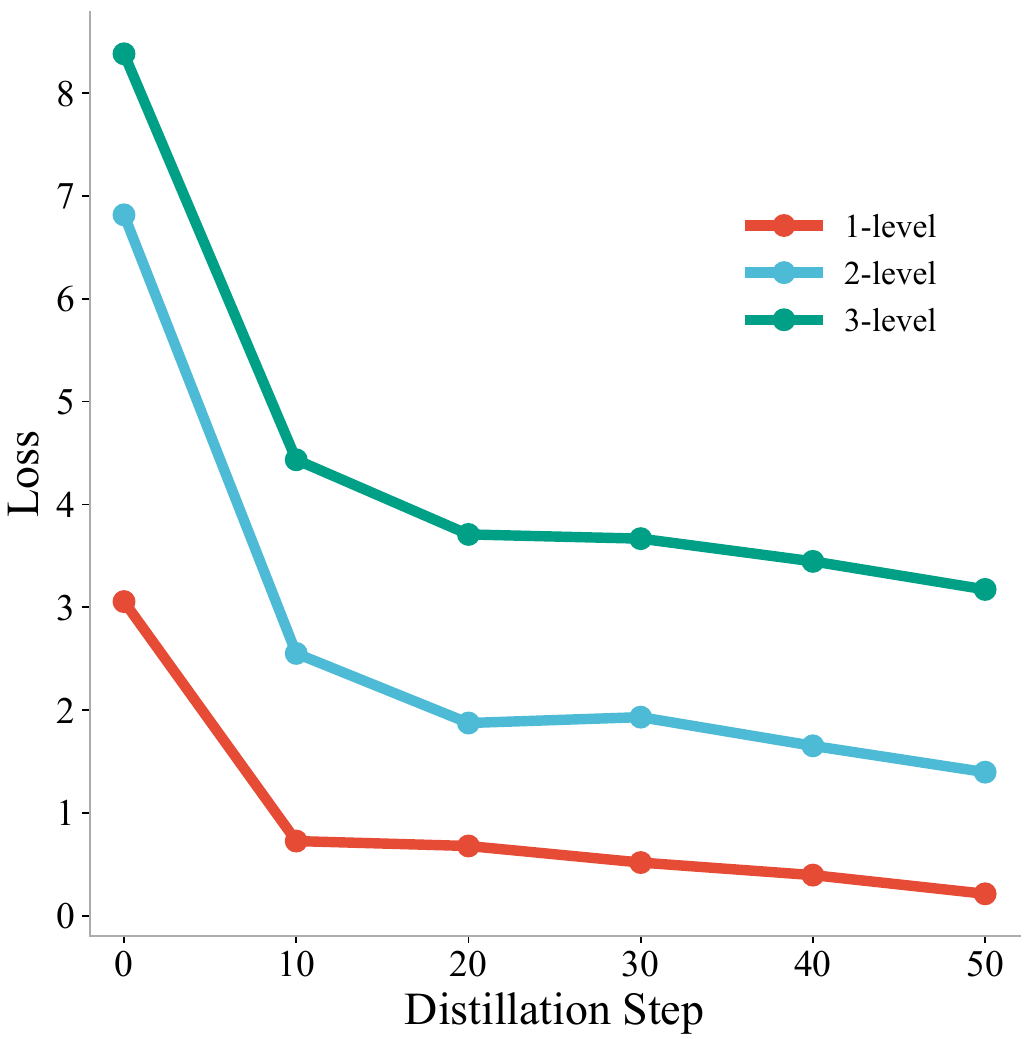}
    }
    \caption{Ablation study on the distillation performance impact of task head’s network parameter size on the foundation model with SPC=1 in the STIZ-Kidney.}
    \label{head}
\end{figure}

we study the distillation performance impact of task head’s network parameter size on the foundation model, the results are shown in Fig~\ref{head}.
It is obvious that under the condition of a fixed foundation model, the larger task head's network parameters and layers embedded in the foundation model, the lower the distillation accuracy and the greater the matching loss. This result indicates that smaller network size in task head can achieve better distillation performance, that is, task head with fewer parameters can more effectively align the gradient or distribution matching information between the expert and student model.

\begin{table}[t]
\centering
\small 
\caption{Ablation study on the performance impact of freezing the foundation model parameter weights during distillation process with SPC=1 in the STIZ-Kidney.}
\label{frozen}
\renewcommand{\arraystretch}{1.3638}
\resizebox{0.98\linewidth}{!}{
\begin{tabular}{cc|cccccc}
\hline & & \multicolumn{6}{c}{\textbf{Foundation model weights frozen}}   \\
Frozen & Step & 0  & 10 & 20  & 30 & 40 & 50    \\
\hline
\multirow{2}{*}{False} & Loss & $2916  $ & $2903  $ & $2894  $ & $2890 $  & $2887 $  & $2883  $   \\
 & Acc & $23.86  $ & $38.17  $ & $40.17  $ & $43.77 $ & $44.73 $  & $45.87  $  \\  \hline

\multirow{2}{*}{True} & Loss & $3.055  $ & $0.726  $ & $0.678  $ & 
$0.517  $ & $0.395 $  & $0.213 $  \\
& Acc & $46.61  $ & $65.29  $ & $67.82  $ & $71.92 $ & $72.37 $  & $73.78  $  \\
\hline
\end{tabular}
}
\end{table}

\begin{figure*}[]
    \centering
    % 第一个子图
    \begin{minipage}[b]{\textwidth}
        \centering
        \includegraphics[width=0.88\textwidth]{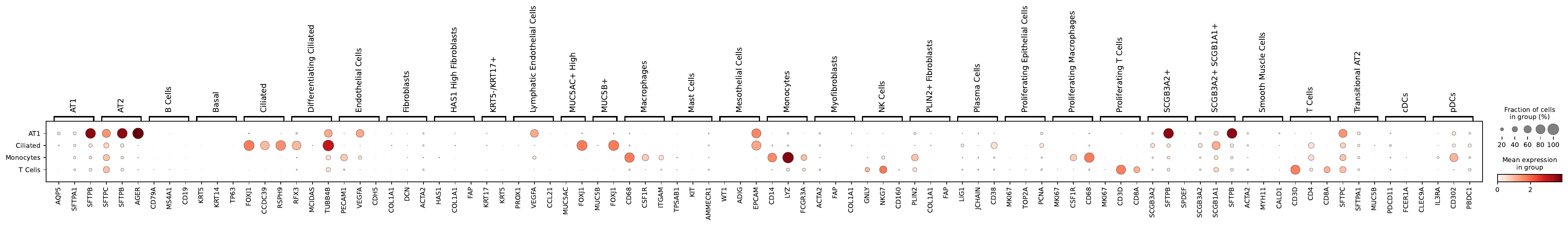}
       % \subcaption{Buffer}
        \label{fig:buffer}
    \end{minipage}
%\vspace{-0.1cm}
    % 第二个子图
    \begin{minipage}[b]{\textwidth}
        \centering
        \includegraphics[width=0.88\textwidth]{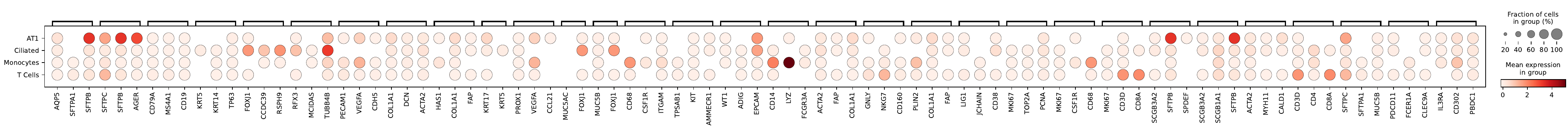}
        %\subcaption{FTD}
        \label{fig:ftd}
    \end{minipage}
%\vspace{-0.5cm}
    % 第三个子图
    \begin{minipage}[b]{\textwidth}
        \centering
        \includegraphics[width=0.88\textwidth]{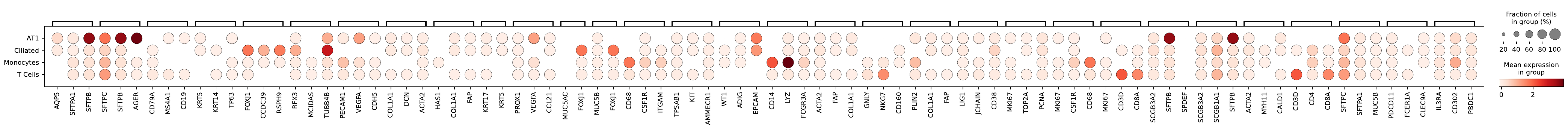}
        %\subcaption{OUR}
        \label{fig:our}
    \end{minipage}

    \caption{Visualize the top three highly variable gene expression values at the gene-level in scRNA-seq dataset, where the scRNA-seq dataset come from the original dataset, gradient-matching synthetic dataset, and synthetic dataset generated by scDD.}
    \label{cc}
\end{figure*}
\begin{figure*}[ht]
        \centering
        \subcaptionbox{Initial synthetic dataset}{
        \includegraphics[width=2.335in]{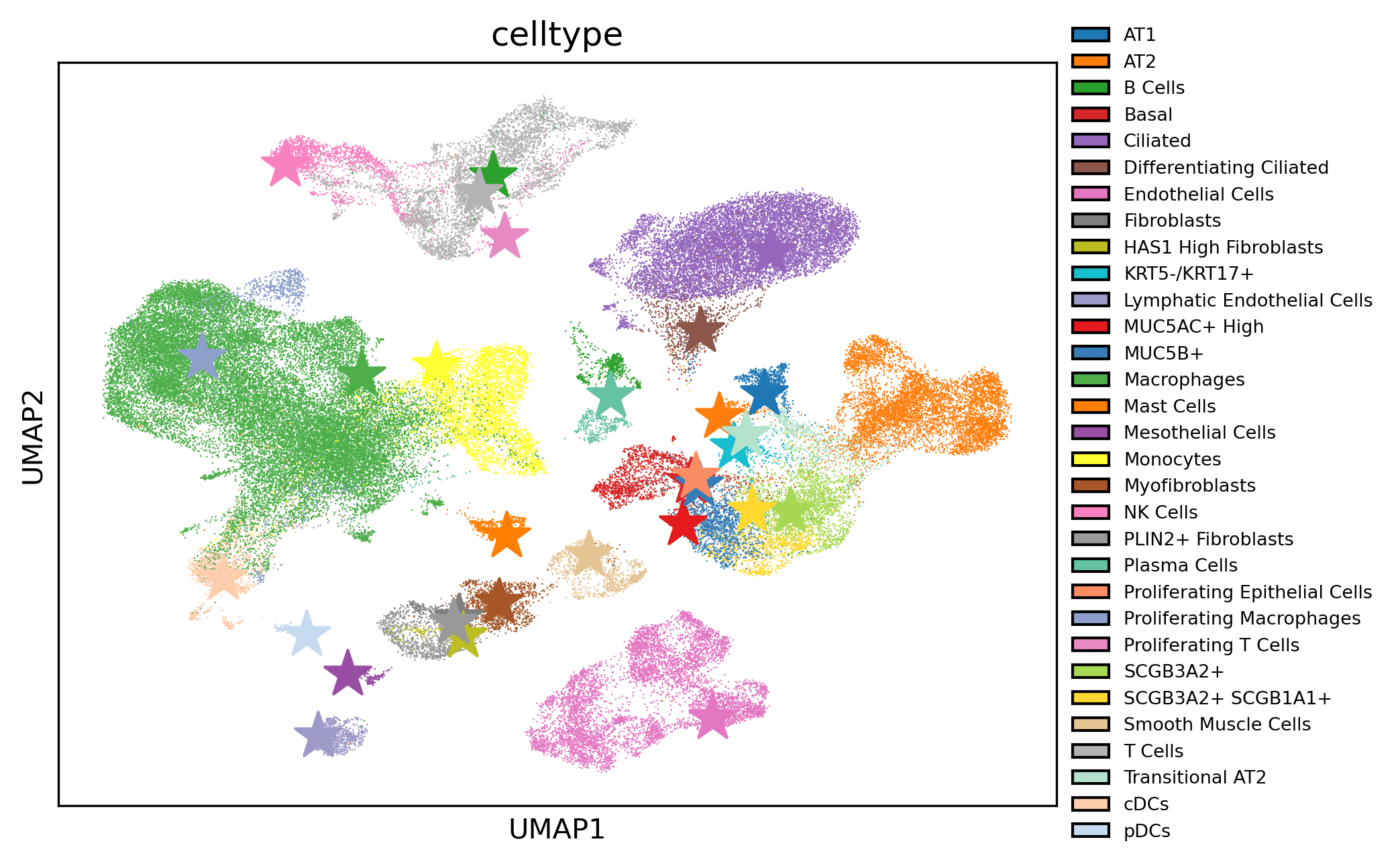}
    }%\hspace{-0.2cm}
        \subcaptionbox{Gradient-matching synthetic dataset}{
        \centering
        \includegraphics[width=2.335in]{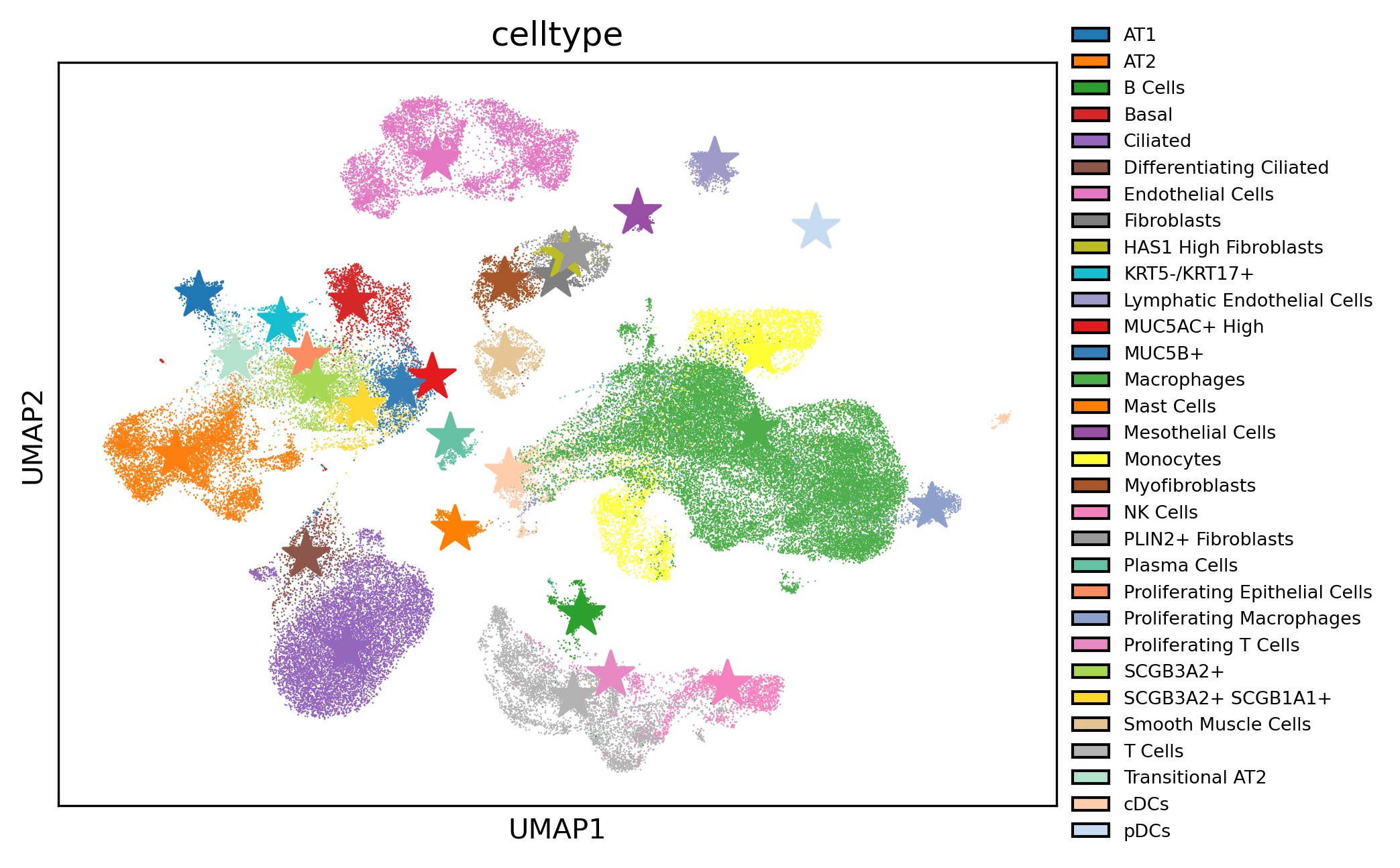}
    }%\hspace{-0.2cm}
        \subcaptionbox{Our scDD synthetic dataset}{
        \includegraphics[width=2.335in]{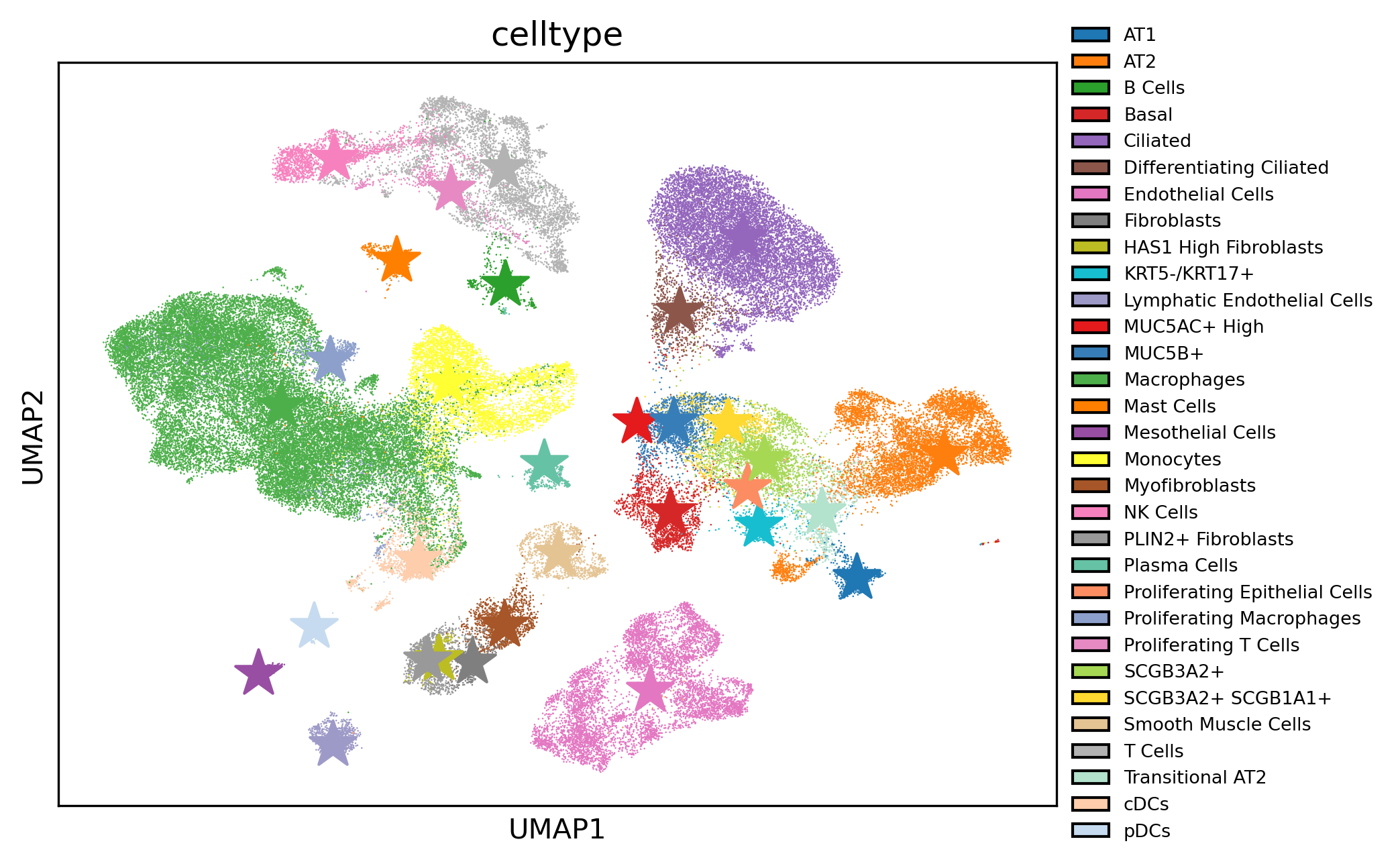}
    }%\hspace{-0.55cm}
    %    \subcaptionbox{overlap}{
    %    \centering
    %    \includegraphics[width=1.71in]{Figs/overlap.pdf}
    %}
    \caption{Visualize the distribution of the synthetic scRNA-seq dataset (Star markers)) within the original scRNA-seq dataset (Point clusters) with different categories using the UMAP in the SCPF-Lung.}
    \label{aa}
\end{figure*}

\subsection{Qualitative Analysis}

\subsubsection{Synthetic dataset distribution visualization}
we visualize the distribution of synthetic dataset (star markers) obtained from different dataset distillation methods within the original dataset (point clusters), and we adopt uniform manifold approximation and projection (UMAP)~\cite{healy2024uniform} to visualize these high-dimensional scRNA-seq dataset. 
UMAP is a manifold learning-based nonlinear dimensionality reduction and clustering visualization method, which balances global and local structural information of high-dimensional data better than t-SNE~\cite{van2008visualizing}, with improved computational and memory efficiency. 
As shown in Fig~\ref{aa}, the synthetic datasets from initialization and gradient-matching do not effectively represent the cell clusters of each category, and the inter-class distances are not sufficiently clear and distinct.
Meanwhile, the synthetic dataset distilled by our method, guided by the SCDG classification conditions and without directly updating the synthetic data, can better represent the multi-class distribution characteristics of the original dataset.

\subsubsection{Gene expression correlations visualization}
\begin{figure}[t]
        \centering
        \subcaptionbox{Original Data}{
        \includegraphics[width=1.68in]{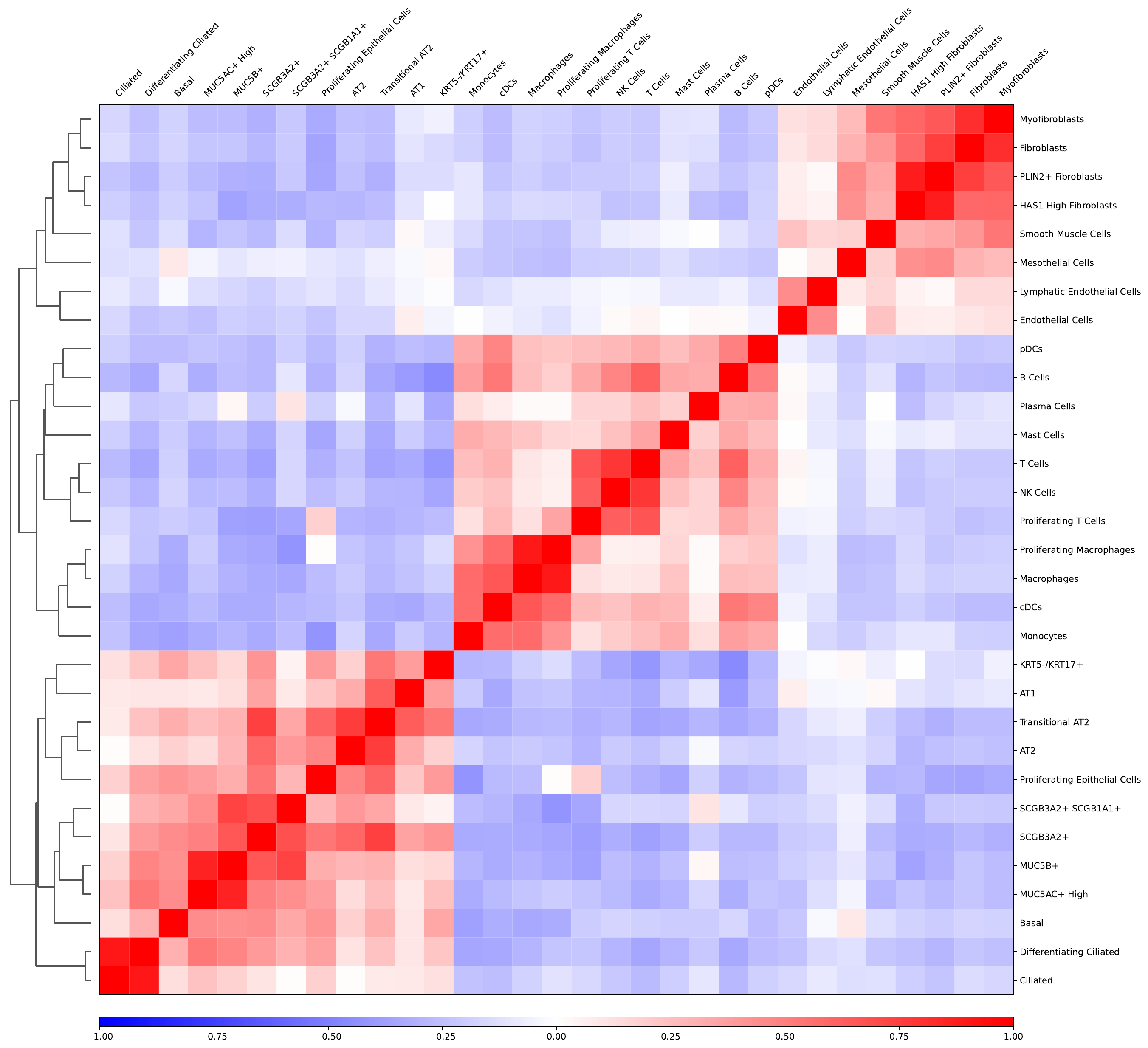}
    }\hspace{-0.3cm}
        \subcaptionbox{SCDG Generated}{
        \centering
        \includegraphics[width=1.68in]{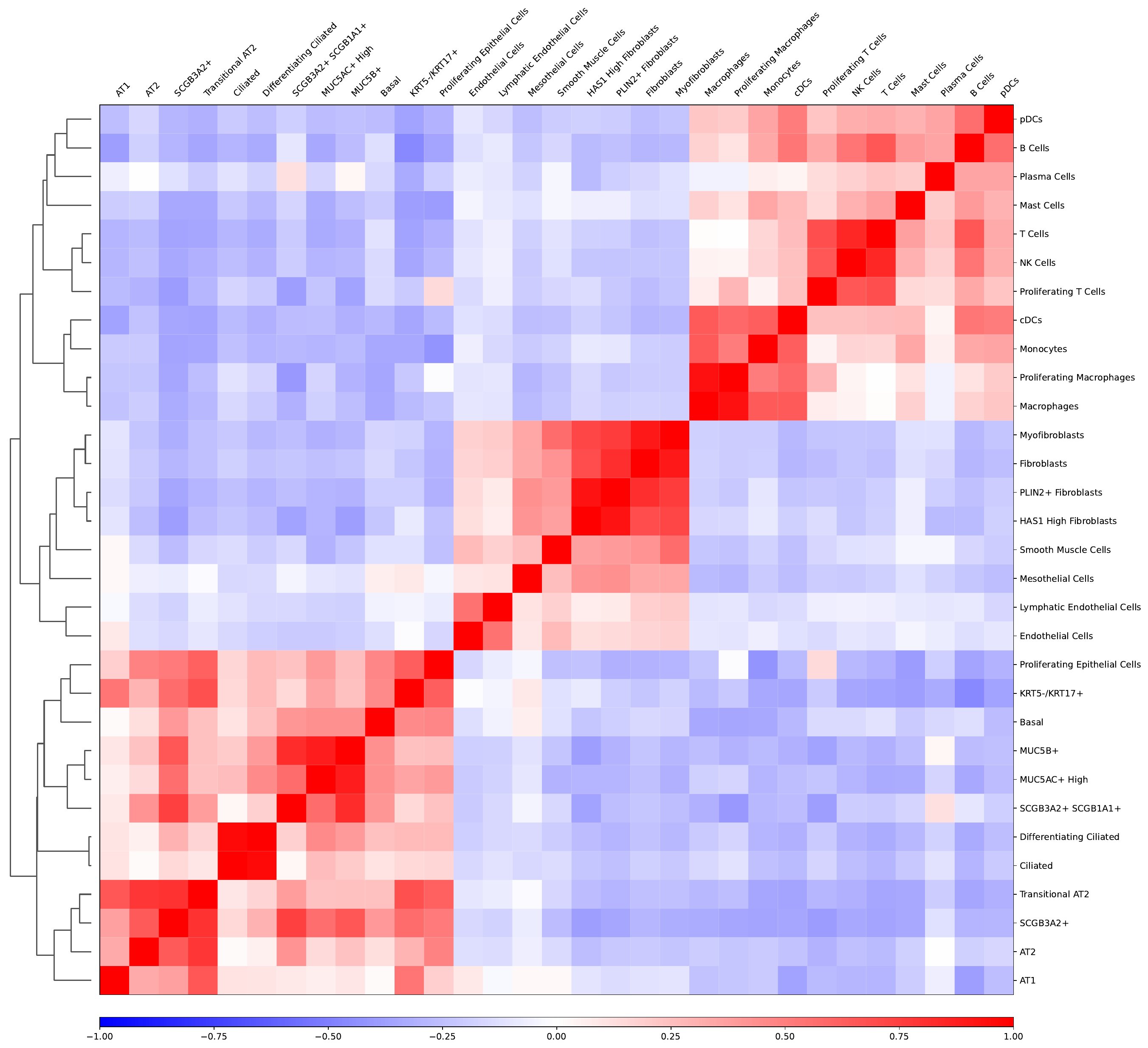}
    }
    \caption{Visualize the gene expression correlations at the cell-level for the original and SCDG generated scRNA-seq dataset.}%\chensays{the font size is too small too see any information}}
    \label{bb}
\end{figure}

we visualize the gene expression correlations at the cell-level for original and SCDG generated scRNA-seq dataset using Scanpy~\cite{li2024analysis}. The matrix-plot can visualize the inter-cluster correlations between cells within a scRNA-seq dataset, which provides an overall reflection of the gene expression values at the cell-level.
As shown in Fig~\ref{bb}, despite the subtle local differences between the synthetic dataset generated by our SCDG and original dataset, it still aligns well with the true biological characteristics of the scRNA-seq data.

\subsubsection{Gene expression values visualization}
we visualize the top three highly variable gene expression values at the gene-level for each single-cell type in the SCPF-Lung dataset, and larger gene expression values are displayed in darker colors. As shown in Fig~\ref{cc}, compared to the original dataset (first row), the synthetic RNA-seq dataset generated by gradient matching (second row) and our method (third row) have disrupted the sparsity of the original gene expressions, and the distribution of these newly added non-zero values is the distilled information from the original dataset. However, due to the effectiveness of our generator and distillation framework, the gene expression distribution of synthetic dataset we distilled is closer to the original dataset.

From the qualitative and quantitative analysis results in distillation, we have summarized the following observations:
(1) The framework of distilling information to the latent-level exhibits better generalization compared to directly updating gene expression values at the data-level. The training results on unseen validation models, where regularization operations such as dropout and activation functions are removed, exhibit performance similar to that of the known distillation model.
(2) To more effectively and accurately align and match the gradient or distribution information between the expert and student model, freezing the foundation model's weights and using a small-scale single-layer task head during the distillation process is the optimal scRNA-seq dataset distillation strategy.
(3) We visualize the gene expression distribution with stronger class-specific characteristics of the synthetic dataset generated by our scDD at both the gene and cell levels, and we also demonstrate that our method can achieve significant performance even at very low compression ratios.
\section{Conclusion}

In this paper, we first propose a latent codes based scRNA-seq dataset distillation framework scDD, which replaces the direct update of gene
expression values at the high-dimensional sparse and non-negative scRNA-seq data-level. Instead, scDD transfers and distills foundation model knowledge and original dataset information into a compact latent space and delivers the final synthetic dataset through a generator. 
Then, we propose a single-step conditional diffusion generator SCDG, which differs from traditional diffusion models that involve multi-step forward and reverse diffusion process. SCDG can perform single-step gradient back-propagation, helping scDD optimize distillation quality and avoid gradient decay caused by multi-step propagation. Meanwhile, SCDG ensures the scRNA-seq data characteristics and inter-class discriminability of the synthetic dataset through flexible conditional control and generation quality assurance.
Finally, we established a comprehensive covered benchmark to evaluate the performance of scRNA-seq dataset distillation in different data analysis tasks, and the empirical results show scDD can significantly improve the distillation performance in the benchmark.

\bibliographystyle{IEEEtran}
\bibliography{reference}

\end{document}